\documentclass[journal,twocolumn]{IEEEtran}

\usepackage{array}
\usepackage{amsmath}
\usepackage{amssymb}
\usepackage{url}
\usepackage[normalem]{ulem} 
\usepackage{mathptmx}
\usepackage{acronym}
\usepackage{booktabs}
\usepackage[T1]{fontenc}
\usepackage{caption}
\usepackage{subcaption}
\usepackage[switch]{lineno}
\usepackage{graphicx}
\usepackage{lipsum} 
\usepackage{fixltx2e} 
\usepackage{color}
\usepackage{xr}
\usepackage[colorlinks,breaklinks]{hyperref} 

\hypersetup{hypertexnames=true,
  linkcolor=blue,anchorcolor=black,citecolor=blue,urlcolor=blue
}
\graphicspath{{imglocal/}{img/}}
\acrodef{ADC}[ADC]{Analog to Digital Converter}
\acrodef{AER}[AER]{Address-Event Representation}
\acrodef{AEX}[AEX]{AER EXtension board}
\acrodef{AE}[AE]{Address-Event}
\acrodef{AFM}[AFM]{Atomic Force Microscope}
\acrodef{AMDA}[AMDA]{AER Motherboard with D/A converters}
\acrodef{ANN}[ANN]{Attractor Neural Network}
\acrodef{API}[API]{Application Programming Interface}
\acrodef{ARM}[ARM]{Advanced RISC Machine}
\acrodef{ASIC}[ASIC]{Application Specific Integrated Circuit}
\acrodef{BCM}[BMC]{Bienenstock-Cooper-Munro}
\acrodef{BD}[BD]{Bundled Data}
\acrodef{BEOL}[BEOL]{Back-end of Line}
\acrodef{BG}[BG]{Bias Generator}
\acrodef{BMI}[BMI]{Brain-Machince Interface}
\acrodef{CAM}[CAM]{Content Addressable Memory}
\acrodef{CAVIAR}[CAVIAR]{Convolution AER Vision Architecture for Real-Time}
\acrodef{CCN}[CCN]{Cooperative and Competitive Network}
\acrodef{CMOL}[CMOL]{``Hybrid CMOS nanoelectronic circuits''}
\acrodef{CMOS}[CMOS]{Complementary Metal-Oxide-Semiconductor}
\acrodef{COTS}[COTS]{Commercial Off-The-Shelf}
\acrodef{CPG}[CPG]{Central Pattern Generator}
\acrodef{CPU}[CPU]{Central Processing Unit}
\acrodef{CV}[CV]{Coefficient of Variation}
\acrodef{DAC}[DAC]{Digital to Analog Converter}
\acrodef{DBN}[DBN]{Deep Belief Network}
\acrodef{DFA}[DFA]{Deterministic Finite Automaton}
\acrodef{DMA}[DMA]{Direct Memory Access}
\acrodef{DNF}[DNF]{Dynamic Neural Field}
\acrodef{DOF}[DOF]{Degrees of Freedom}
\acrodef{DPE}[DPE]{Dynamic Parameter Estimation}
\acrodef{DPI}[DPI]{Differential Pair Integrator}
\acrodef{DRAM}[DRAM]{Dynamic Random Access Memory}
\acrodef{DR}[DR]{Dual Rail}
\acrodef{DSP}[DSP]{Digital Signal Processor}
\acrodef{DVS}[DVS]{Dynamic Vision Sensor}
\acrodef{EBL}[EBL]{Electron Beam Lithography}
\acrodef{EDVAC}[EDVAC]{Electronic Discrete Variable Automatic Computer}
\acrodef{EIN}[EIN]{Excitatory-Inhibitory Network}
\acrodef{EM}[EM]{Expectation Maximization}
\acrodef{EPSC}[EPSC]{Excitatory Post-Synaptic Current}
\acrodef{EPSP}[EPSP]{Excitatory Post-Synaptic Potential}
\acrodef{FDSOI}[FDSOI]{Fully-Depleted Silicon on Insulator}
\acrodef{FET}[FET]{Field-Effect Transistor}
\acrodef{FFT}[FFT]{Fast Fourier Transform}
\acrodef{FI}[F-I]{Frequency-Current}
\acrodef{FPGA}[FPGA]{Field Programmable Gate Array}
\acrodef{FSA}[FSA]{Finite State Automaton}
\acrodef{FSM}[FSM]{Finite State Machine}
\acrodef{GOPS}[GOPS]{Giga-Operations per Second}
\acrodef{GPU}[GPU]{Graphical Processing Unit}
\acrodef{HAL}[HAL]{Hardware Abstraction Layer}
\acrodef{HH}[H\&H]{Hodgkin \& Huxley}
\acrodef{HMM}[HMM]{Hidden Markov Model}
\acrodef{HRS}[HRS]{High-Resistive State}
\acrodef{HR}[HR]{Human Readable}
\acrodef{HW}[HW]{Hardware}
\acrodef{ICT}[ICT]{Information and Communication Technology}
\acrodef{IC}[IC]{Integrated Circuit}
\acrodef{IF2DWTA}[IF2DWTA]{Integrate \& Fire 2--Dimensional WTA}
\acrodef{IFSLWTA}[IFSLWTA]{Integrate \& Fire Stop Learning WTA}
\acrodef{IF}[I\&F]{Integrate-and-Fire}
\acrodef{IMU}[IMU]{Inertial Measurement Unit}
\acrodef{INCF}[INCF]{International Neuroinformatics Coordinating Facility}
\acrodef{INI}[INI]{Institute of Neuroinformatics}
\acrodef{IO}[I/O]{Input/Output}
\acrodef{IPSC}[IPSC]{Inhibitory Post-Synaptic Current}
\acrodef{IPSP}[IPSP]{Inhibitory Post-Synaptic Potential}
\acrodef{IP}[IP]{Intellectual Property}
\acrodef{ISI}[ISI]{Inter-Spike Interval}
\acrodef{JFLAP}[JFLAP]{Java - Formal Languages and Automata Package}
\acrodef{LFP}[LFP]{Local Field Potential}
\acrodef{LNA}[LNA]{Low-Noise Amplifier}
\acrodef{LPF}[LPF]{Low-Pass Filter}
\acrodef{LRS}[LRS]{Low-Resistive State}
\acrodef{LSM}[LSM]{Liquid State Machine}
\acrodef{LTD}[LTD]{Long Term Depression}
\acrodef{LTI}[LTI]{Linear Time-Invariant}
\acrodef{LTP}[LTP]{Long Term Potentiation}
\acrodef{LTU}[LTU]{Linear Threshold Unit}
\acrodef{LUT}[LUT]{Look-Up Table}
\acrodef{MCMC}[MCMC]{Markov-Chain Monte Carlo}
\acrodef{MEMS}[MEMS]{Micro Electro Mechanical System}
\acrodef{MIM}[MIM]{Metal Insulator Metal}
\acrodef{MI}[MI]{Mutual Information}
\acrodef{MLP}[MLP]{Multy-Layer Perceptron}
\acrodef{MOSCAP}[MOSCAP]{Metal Oxide Semiconductor Capacitor}
\acrodef{MOSFET}[MOSFET]{Metal Oxide Semiconductor Field-Effect Transistor}
\acrodef{MOS}[MOS]{Metal Oxide Semiconductor}
\acrodef{NDFSM}[NDFSM]{Non-deterministic Finite State Machine} 
\acrodef{ND}[ND]{Noise-Driven}
\acrodef{NEF}[NEF]{Neural Engineering Framework}
\acrodef{NHML}[NHML]{Neuromorphic Hardware Mark-up Language}
\acrodef{NIL}[NIL]{Nano-Imprint Lithography}
\acrodef{NMDA}[NMDA]{N-Methyl-D-Aspartate}
\acrodef{NME}[NE]{Neuromorphic Engineering}
\acrodef{OTA}[OTA]{Operational Transconductance Amplifier}
\acrodef{PCB}[PCB]{Printed Circuit Board}
\acrodef{PSC}[PSC]{Post-Synaptic Current}
\acrodef{PSTH}[PSTH]{Peri-Stimulus Time Histogram}
\acrodef{RAM}[RAM]{Random Access Memory}
\acrodefplural{RAM}[RAMs]{Random Access Memories}
\acrodef{RMSE}[RMSE]{Root Mean Squared-Error}
\acrodef{RMS}[RMS]{Root Mean Squared}
\acrodef{RNN}[RNN]{Recurrent Neural Network}
\acrodef{RRAM}[RRAM]{Resistive Random Access Memory}
\acrodefplural{RRAM}[R-RAMs]{Resistive Random Access Memories}
\acrodef{SAC}[SAC]{Selective Attention Chip}
\acrodef{SCX}[SCX]{Silicon CorteX}
\acrodef{SD}[SD]{Signal-Driven}
\acrodef{SEM}[SEM]{Spike-based Expectation Maximization}
\acrodef{SOI}[SOI]{Silicon on Insulator}
\acrodef{SOC}[SOC]{System-On-Chip}
\acrodef{SRAM}[SRAM]{Static Random Access Memory}
\acrodef{STDP}[STDP]{Spike-Timing Dependent Plasticity}
\acrodef{STD}[STD]{Short-Term Depression}
\acrodef{STP}[STP]{Short-Term Plasticity}
\acrodef{STT}[STT]{Spin-Transfer Torque}
\acrodef{STT-MRAM}[STT-MRAM]{Spin-Transfer Torque Magnetic Random Access Memory}
\acrodefplural{STT-MRAM}[STT-MRAMs]{Spin-Transfer Torque Magnetic Random Access Memories}
\acrodef{SW}[SW]{Software}
\acrodef{TFT}[TFT]{Thin Film Transistor}
\acrodef{USB}[USB]{Universal Serial Bus}
\acrodef{VHDL}[VHDL]{VHSIC Hardware Description Language}
\acrodef{VLSI}[VLSI]{Very Large Scale Integration}
\acrodef{VOR}[VOR]{Vestibulo-Ocular Reflex}
\acrodef{WTA}[WTA]{Winner-Take-All}
\acrodef{XML}[XML]{eXtensible Mark-up Language}
\acrodef{divmod3}[DIVMOD3]{divisibility of a number by 3}
\acrodef{hWTA}[hWTA]{Hard Winner-Take-All}
\acrodef{sWTA}[sWTA]{soft Winner-Take-All}

\acrodef{NP}[ROLLS neuromorphic processor]{Reconfigurable On-line Learning Spiking Neuromorphic Processor}

\definecolor{gray}{rgb}{0.75,0.75,0.75}
\newcommand{\revised}[3]{#2}
\newcommand{\revisednolines}[1]{#1}

\begin{document}



%
\title{Memory and information processing in neuromorphic systems}
\author{Giacomo~Indiveri~\IEEEmembership{Senior Member,~IEEE} and 
Shih-Chii~Liu~\IEEEmembership{Senior Member,~IEEE}
\thanks{G. Indiveri and S.-C. Liu are with the Institute of Neuroinformatics, University of Zurich and ETH Zurich, Switzerland}
\thanks{Manuscript received Month DD, YEAR; revised MONTH DD, YEAR.}}

\markboth{Proceedings of the IEEE,~Vol.~x, No.~x, June~2015}%
{Indiveri:  Memory and information processing in neuromorphic systems}
%

\maketitle

\begin{abstract}

A striking difference between brain-inspired neuromorphic processors and current von Neumann processors architectures is the way in which memory and processing is organized. As Information and Communication Technologies continue to address the need for increased computational power through the increase of cores within a digital processor, neuromorphic engineers and scientists can complement this need by building processor architectures where memory is distributed with the processing. 
In this paper we present a survey of brain-inspired processor architectures that support models of cortical networks and deep neural networks. These architectures range from serial clocked implementations of multi-neuron systems to massively parallel asynchronous ones and from purely digital systems to mixed analog/digital systems which implement more biological-like models of neurons and synapses together with a suite of adaptation and learning mechanisms analogous to the ones found in biological nervous systems. We describe the advantages of the different approaches being pursued and present the challenges that need to be addressed for building artificial neural processing systems that can display the richness of behaviors seen in biological systems.
\end{abstract}



%
\IEEEpeerreviewmaketitle

\section{Introduction} %
\label{sec:introduction}

\begin{figure}
  \begin{subfigure}{0.45\textwidth}
    \centering
    \includegraphics[width=\textwidth]{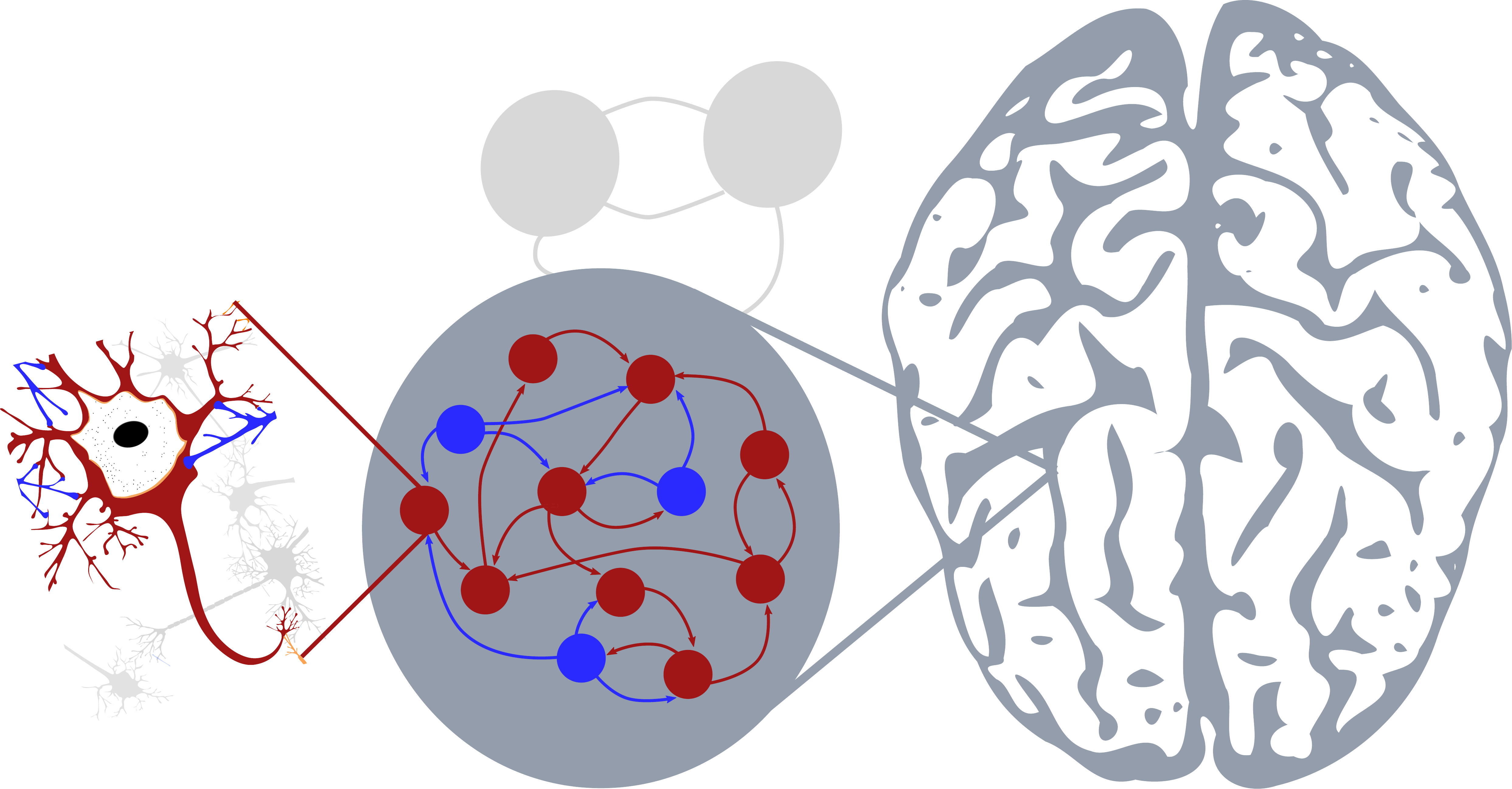}
    \subcaption{} 
    \label{fig:memory-brain}
  \end{subfigure}\\
  \begin{subfigure}{0.45\textwidth}
    \centering
    \includegraphics[width=\textwidth]{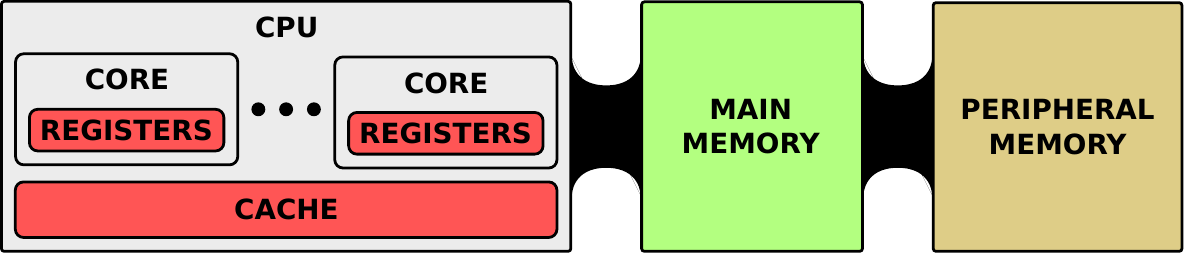}
    \subcaption{} 
    \label{fig:memory-pc}
  \end{subfigure}
  \caption{Memory hierarchies in brains and computers. In brains, (\subref{fig:memory-brain}) Neurons and synapses are the fundamental elements of both neural computation and memory formation. Multiple excitatory and inhibitory neurons, embedded in recurrent canonical microcircuits, form basic computational primitives that can carry out state-dependent sensory processing and computation. Multiple clusters of recurrent networks are coupled together via long-distant connections to implement sensory fusion, inference, and symbolic manipulation.  In computers, (\subref{fig:memory-pc}) \aclp{CPU} containing multiple cores are connected to both main memory and peripheral memory blocks. Each core comprises a micro-processor and local memory (e.g., local registers and Level-1 cache). All cores typically share access to another block of fast memory integrated on the same chip (e.g., Level-2 cache). The main memory block is the primary storage area, typically larger than the memory blocks inside the \ac{CPU}, but requiring longer access times. The peripheral memory block requires even longer access rates, but can store significantly larger amounts of data.}
  \label{fig:memory-hierarchies}
\end{figure}

Neuromorphic information processing systems consist of electronic circuits and devices built using design principles that are based on those of biological nervous systems~\cite{Mead90,Douglas_etal95a,Indiveri_Horiuchi11,Chicca_etal14}.  The circuits are typically designed using mixed-mode analog/digital \ac{CMOS} transistors and fabricated using standard \ac{VLSI} processes. Similar to the biological systems that they model, neuromorphic systems process information using energy-efficient asynchronous, event-driven, methods~\cite{Liu_etal14}. They are often adaptive, fault-tolerant, and can be flexibly configured to display complex behaviors by combining multiple instances of simpler elements. The most striking difference between neuromorphic systems and conventional information processing systems is in their use of memory structures. While computing systems based on the classical von Neumann architecture have one or more central processing units physically separated from the main memory areas, both biological and artificial neural processing systems are characterized by co-localized memory and computation (see Fig.~\ref{fig:memory-hierarchies}): the synapses of the neural network implement at the same time memory storage as well as complex non-linear operators used to perform collective and distributed computation. Given that memory-related constraints, such as size, access latency and throughput, represent one of the major performance bottlenecks in conventional computing architectures~\cite{Backus78}, and given the clear ability of biological nervous systems to perform robust computation, using memory and computing elements that are slow, in-homogeneous, stochastic and faulty~\cite{Habenschuss_etal13,Maass14}, neuromorphic brain inspired computing paradigms offer an attractive solution for implementing alternative non von Neumann architectures, using advanced and emerging technologies.

This neuromorphic engineering approach, originally proposed in the late eighties~\cite{Mead89} and pursued throughout the nineties~\cite{Douglas_etal95a, Fragniere_etal97, Horiuchi_Koch99} and early 2000s~\cite{Liu_etal02a, Boahen05, Sarpeshkar06} by a small number of research labs worldwide, is now being adopted by an increasing number of both academic and industrial research groups. \revised{memristors-neuromorphic}{In particular, there have been many recent publications describing the use of new materials and nano-technologies for building nano-scale devices that can emulate some of the properties observed in biological synapses~\cite{Jo_etal10,Kim_etal11,Yu_etal13,Suri_etal13,Kim_etal15,Saighi_etal15}. At the network and system level}{Recently} remarkable brain-inspired electronic multi-neuron computing platforms have been developed to implement alternative computing paradigms for solving pattern recognition and machine learning tasks~\cite{Merolla_etal14a,Furber_etal14} and for speeding-up the simulation of computational neuroscience models~\cite{Furber_etal14, Pfeil_etal13}. These \revised{latter}{latter}{} approaches however are only loosely inspired by  biological neural processing systems, and are constrained by both precision requirements (e.g., with digital circuits, to guarantee bit-precise equivalence with software simulations, and with analog circuits, to implement as faithfully as possible the equations and specifications provided by the neuro-scientists), and bandwidth requirements (e.g., to speed up the simulations by two or three orders of magnitude, or to guarantee that all transmitted signals reach their destinations within some clock cycle duration). 

An alternative strategy is to forgo these constraints and emulate biology much more closely by \revised{new-materials}{developing new materials and devices, and by}{} designing electronic circuits that exploit \revised{device-physics}{their device physics}{the physics of the silicon medium} to reproduce the bio-physics of real synapses, neurons, and other neural structures~\cite{Mead89,Jo_etal10,Rajendran_etal13,Benjamin_etal14,Chicca_etal14}. \revised{CMOS}{In \ac{CMOS}, this}{This} can be achieved by using \acp{FET} operated in the analog ``weak-inversion'' or ``sub-threshold'' domain~\cite{Liu_etal02a}, which naturally exhibit exponential relationships in their transfer functions, similar for example, to the exponential dependencies observed in the conductance of Sodium and Potassium channels of biological neurons~\cite{Hodgkin_Huxley52}. In the sub-threshold domain, the main mechanism of carrier transport is diffusion and many of the computational operators used in neural systems (such as exponentiation, thresholding, and amplification) can be implemented using circuits consisting of only a few transistors, sometimes only one.  Therefore, sub-threshold analog circuits require far fewer transistors than digital for emulating certain properties of neural systems. However these circuits tend to be slow, inhomogeneous, and imprecise.  To achieve fast, robust, and reliable information processing in neuromorphic systems designed following this approach, it is necessary to adopt computational strategies that are analogous to the ones found in nature: for fast processing, low latency, and quick response times these strategies include using massively parallel arrays of processing elements that are asynchronous, real-time, and data- or event-driven (e.g., by responding to or producing spikes). For robust and reliable processing, crucial strategies include both the co-localization of memory and computation, and the use of adaptation and plasticity mechanisms that endow the system with stabilizing and learning properties. 

In this paper we will present an overview of current approaches that implement memory and information processing in neuromorphic systems. The systems range from implementations of neural networks using conventional von Neumann architectures, to custom hardware implementations that have co-localized memory and computation elements, but which are only loosely inspired by biology, to neuromorphic architectures which implement biologically plausible neural dynamics and realistic plasticity mechanisms, merging both computation and memory storage within the same circuits. We will highlight the advantages and disadvantages of these approaches, pointing out which application domains are best suited to them, and describe the conditions where they can best exploit the properties of new materials and devices, such as \revised{no-memristors}{oxide-based resistive memories}{memristors} and spin-Field Effect Transistors (spin-FETs). 

\revised{applications}{
\subsection{Application areas for neuromorphic systems}
\label{sec:applications}

Although the origin of the field of neuromorphic engineering can be traced back to the late '80s, this field is still relatively young when considering the amount of man-power that has been invested in it. Therefore there are not many well-established products and applications in the market that exploit neuromorphic technology to its full potential yet. However, it has been argued that there are several areas in which  neuromorphic systems offer significant advantages over  conventional computers~\cite{Boahen05,Sarpeshkar06,OConnor_etal13,Chicca_etal14,Liu_etal14}, such as that of sensory processing~\cite{Liu_Delbruck10} or ``autonomous  systems''. An autonomous system can be a simple one, such as a sensory processing system based on environmental sensors or bio-sensors; or an intermediate-complexity one, such as a \ac{BMI} making one or two bit decisions based on the real-time on-line processing of small numbers of signals, sensed continuously from the environment~\cite{Corradi_Indiveri15}; or a complex one, such as a humanoid robot making  decisions and producing behaviors based on the outcome of sophisticated auditory or visual processing~\cite{Bartolozzi_etal11c}. These types of autonomous systems can greatly benefit from the extremely compact and low-power features of the neuromorphic hardware technology~\cite{Mead90} and can take advantage of the neural style of computation that the neuromorphic hardware substrate supports, to develop new computing paradigms that are better suited to unreliable sensory signals in uncontrolled environments. 
 
Another application domain where dedicated neural processing hardware systems are being used to complement or even to replace conventional computers is that of custom accelerated simulation engines for large-scale neural modeling~\cite{Furber_etal14,Schemmel_etal10}, or of very large-scale spiking neural networks applied to machine learning problems~\cite{Merolla_etal14a}. Also in this application area, the low-power features of the dedicated neuromorphic hardware implementations typically outperform those of general purpose computing architectures used for the same purposes. In Section~\ref{sec:electronic-systems}, we will describe deep network architectures currently used for machine learning benchmarks and describe examples of neuromorphic hardware network implementations that have been proposed to replace conventional computers used in simulating these architectures. In Section~\ref{sec:cust-memory-optim} we will describe examples of hardware systems that have been proposed to implement efficient simulations of large-scale neural models. Section~\ref{sec:learning} will present an overview of the adaptation, learning, and memory mechanisms adopted by the nervous system to carry out computation, and Section~\ref{sec:neur-proc} will present an example of a neuromorphic processor that implements such mechanisms and that can be used to endow autonomous behaving systems with learning abilities for adapting to changes in the environment and interacting with it in real-time.
}{}

\section{From \acp{CPU} to deep neural networks}
\label{sec:electronic-systems}

\begin{figure}
  \begin{subfigure}{0.225\textwidth}
    \centering
    \includegraphics{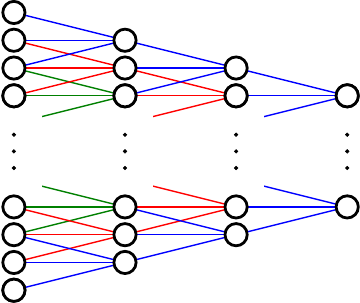}
    \subcaption{} 
    \label{fig:convolutional}
  \end{subfigure}
  \begin{subfigure}{0.225\textwidth}
    \centering
    \includegraphics{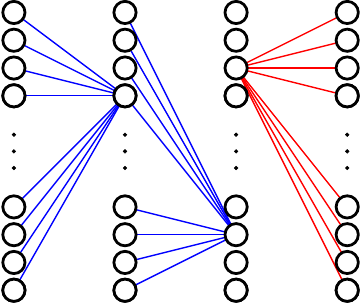}
    \subcaption{} 
    \label{fig:deep}
  \end{subfigure}
  \caption{Multi-layer neural networks. (\subref{fig:convolutional}) Hierarchical convolutional network with feed-forward connections; (\subref{fig:deep}) Deep neural network with all-to-all connections. Converging connections are only shown for two representative neurons (i.e., with large fan-in) in the bottom two layers, while an example of a neuron projecting to multiple targets (i.e., a neuron with large fan-out) is shown only in the second-to-last layer. Connections in (\subref{fig:deep}) can be both feed-forward and feed-back.}
  \label{fig:multi-layer}
\end{figure}

Conventional computers based on the von Neumann architecture typically have one or more \acp{CPU} physically separated from the program and data memory elements (see Fig.~\ref{fig:memory-pc}). The \acp{CPU} access both data and program memory using the same shared resources. Since there is a limited throughput between the processors and the memory, and since processors speeds are much higher than memory access ones, \acp{CPU} spend most of their time idle. This famous von Neumann bottleneck problem~\cite{Backus78} can be alleviated by adding hierarchical memory structures inside the \acp{CPU} to cache frequently used data, or by shifting the computational paradigm from serial to parallel. Driven mainly by performance improvement demands, we have been witnessing both strategies in recent years~\cite{Cavin_etal08}. However, if one takes into account  energy consumption constraints, increasing the size or improving the performance of cache memory is not an option. The energy consumption of cache memory is linearly proportional to its size~\cite{Kamble_Ghose97}. The alternative strategy is therefore to increase the number of parallel computing elements in the system. Amdahl's law~\cite{Amdahl67} has often been used to evaluate the performance gains of parallel processing in multi-processor von Neumann architectures~\cite{Hill_Marty08}. In~\cite{Cassidy_Andreou12} the authors demonstrate how performance gained from parallelism is more energy efficient than performance gained from advanced memory optimization and micro-architectural techniques, by deriving a generalized form of Amdahl's law which takes into account communication, energy and delay factors. A first step toward the implementation of  massively parallel neural processing systems is to explore the use of \acp{GPU},  which typically combine hundreds of parallel cores with high memory bandwidths. Indeed several neural simulation tools have been proposed for this purpose~\cite{Brette_Goodman12,Ciresan_etal10,Nageswaran_etal09}. However, as demonstrated with the cost function derived from the generalized Amdahl's law in~\cite{Cassidy_Andreou12}, even conventional \ac{GPU} architectures are not optimally suited to running these spiking neural network simulations, when energy consumption is factored in. This is why new custom hardware accelerated solutions, with  memory access and routing schemes optimized specifically for neural networks started to emerge. \revised{CPUDNN1}{In particular, a new set of digital architectures have been proposed for implementing ``deep'' multi-layer neural  networks including full custom \ac{CMOS} solutions~\cite{Conti_Benini15,Camunas-Mesa_etal12,Pham_etal12} and solutions based on \ac{FPGA} devices~\cite{Neil_Liu14,Gokhale_etal14}.} {In particular, a new set of custom digital information processing architectures are being proposed for implementing ``deep'' multi-layer neural  networks~\cite{Conti_Benini15,Neil_Liu14,Camunas-Mesa_etal12,Farabet_etal11}.}

\subsection{Deep Networks}
\label{sec:deep-networks}

Deep networks are neural networks composed of many layers of neurons (see Fig.~\ref{fig:multi-layer}). They are currently the network architecture of choice in the machine learning community for solving a wide range of classification problems, and have shown state-of-art performance in various benchmarks tasks such as digit recognition~\cite{Schmidhuber15}. They include convolutional networks which are being explored intensively within the neuromorphic community for visual processing tasks~\cite{Farabet_etal13}. 

\subsection{Convolutional Networks}
\label{sec:fpga-convnet}

Convolutional networks, consist of a multi-layer feed-forward network architecture in which neurons in one layer receive inputs from multiple neurons in the previous layer and produce an output which is a thresholded or sigmoidal function of the weighted sum of its inputs  (see Fig.~\ref{fig:multi-layer} (a)). The connectivity pattern between the units of one layer and the neuron of the subsequent layer, responsible for the weighted sum operation forms the convolution kernel. Each layer typically has one or a small number of convolution kernels that map the activity of a set of neurons from one layer to the target neuron of the subsequent layer. These networks were originally inspired by the structure of the visual system in mammals~\cite{Hubel_Wiesel62,Fukushima13}, and were used extensively for image processing and machine vision tasks~\cite{Marr82,Le-Cun_etal98,Riesenhuber_Poggio99}. They are typically implemented on \acp{CPU} and \acp{GPU} which consume a substantial amount of power. In recent years, alternate dedicated \ac{SOC} solutions and \ac{FPGA} platforms have been used to implement these networks for increasing their performance while decreasing their power consumption. 
Two main approaches are being pursued, depending on the readout scheme of the vision sensor: a \emph{frame-based} approach, which uses inputs from conventional frame-based cameras, and an \emph{event-driven} one, which uses inputs from event-driven retina-like vision sensors~\cite{Brandli_etal14, Lichtsteiner_etal08,Serrano-Gotarredona_Linares-Barranco13,Posch_etal11}.

\subsubsection{Frame-Based Solution}
\label{sec:frame-based-solution}

\begin{figure}
\centering
\includegraphics[width=0.475\textwidth]{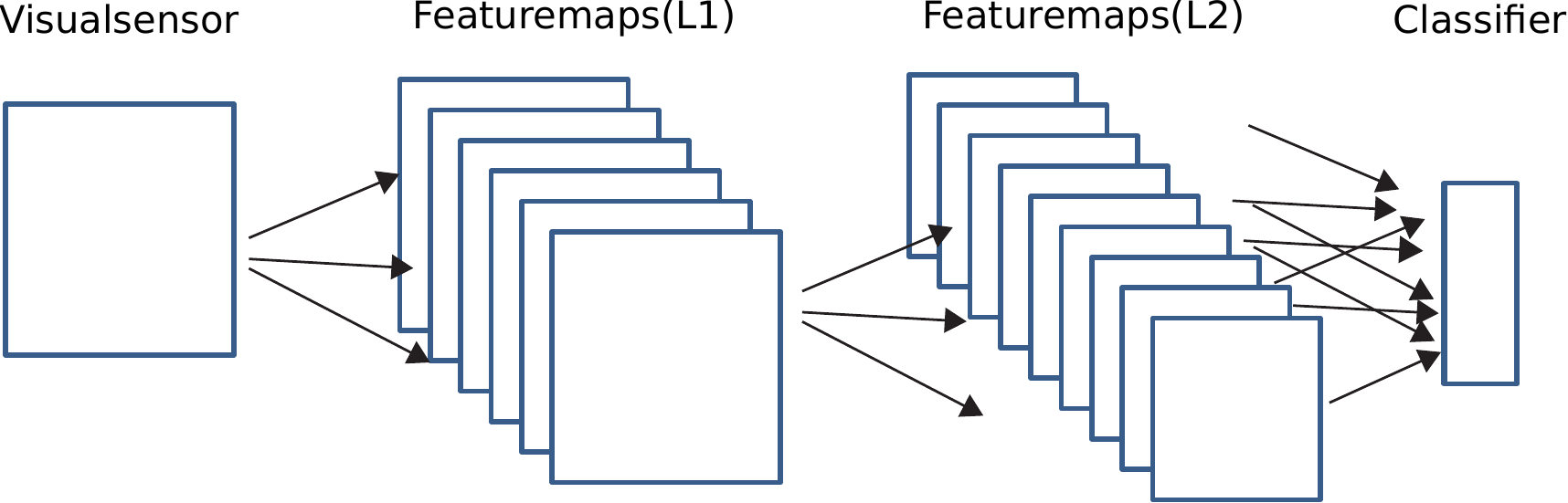}
\caption{Example architecture of a convolutional multi-layered network.}
\label{fig:convnetblock}
\end{figure}

An example of a method proposed for implementing scalable multi-layered synthetic vision systems based on  a dataflow architecture such as the one shown in Fig.~\ref{fig:convnetblock} is the ``neuFlow'' system~\cite{Farabet_etal11}.
The dataflow architecture of this system relies on a 2D grid  of Processing Tiles (PTs) where each PT has a bank of operators such as a multiply, divide, add, subtract, and a max; a multiplexer based on-chip router; and a configurable memory mapper block. The architecture is designed to process large streams of data in parallel.
It uses a Smart \ac{DMA} block which interfaces with off-chip memory and provides asynchronous data transfers with priority management. The \ac{DMA} can be configured to read or write a particular chunk of data, and sends its status to another  block called the Flow-CPU. The Flow-CPU works as a central Control Unit that can reconfigure the computing grid and the Smart \ac{DMA} at runtime. The configuration data from Flow-CPU placed on a Runtime Configuration Bus re-configures most aspects of the grid at runtime.
A full-blown compiler, dubbed ``LuaFlow'', takes sequential tree-like or flow-graph descriptions of algorithms and parses them to extract different levels of parallelism. With the implementation of this architecture on a Xilinx Virtex 6 ML605 \ac{FPGA}, the authors demonstrate the segmentation and classification of a street scene using a 4 layered network running at 12\,frames/second. The same  architecture was also implemented in a custom 45\,nm \ac{SOI} process \ac{ASIC} chip, which was predicted to have, by software simulations,  a peak performance of 320\,\ac{GOPS}  with a 0.6\,W power budget~\cite{Pham_etal12}. In comparison, the neuFlow architecture implemented on a standard Xilinx Virtex 6 ML605 \ac{FPGA}  has a peak performance of 16\,\ac{GOPS} with 10\,W of power consumption.

To cope with networks of larger numbers of layers and with their related  memory bandwidth requirements, a scalable low power system called ``nn-X'' was presented in~\cite{Gokhale_etal14, Dundar_etal14}. It comprises a host processor, a coprocessor and external memory. The coprocessor includes an array of processing elements called collections, a memory router, and a configuration bus. Each collection contains a convolution engine, a pooling module, and a nonlinear operator. The memory router routes the independent data stream to the collections and allows nn-X to have access to multiple memory buffers at the same time. The nn-X system as prototyped on the Xilinx ZC706 platform, has eight collections, each with a 10$\times$10 convolution engine, and has a measured performance of 200\,\ac{GOPS} while consuming 4\,W.

\subsubsection{Event-Driven Solution}
\label{sec:event-driv-solut}

\revised{event-cnn}{The implementation of event- or spike-based convolutional network chips was first investigated in the spike-based multi-chip neuromorphic vision system "CAVIAR"\footnote{\revisednolines{The CAVIAR acronym stands for ``Convolution Address-Event Representation (AER) Vision Architecture for Real-Time''. This was a four year project funded by the European Union under the FP5-IST program in June 2002, within the ``Lifelike perception systems'' subprogram. Its main objective was to develop a bio-inspired multi-chip multi-layer hierarchical sensing/processing/actuation system where the chips communicate using an AER infrastructure.}} ~\cite{Serrano-Gotarredona_etal99}. This system consists of a front-end retina chip, a set of spiking convolution chips, a winner-take-all chip, and a set of learning chips. While the first convolutions chips in CAVIAR were designed using mixed-signal analog/digital circuits, a custom digital version with an array of 32$\times$32 pixels was later implemented in a standard 0.35\,$\mu$m \ac{CMOS} process, containing an arbitrary programmable kernel size of up to 32$\times$32~\cite{Camunas-Mesa_etal11}. Extending these chips to implement large-scale networks would require an infrastructure for routing spikes between multiple convolutional chips. In CAVIAR this was done using an asynchronous communication protocol based on the \ac{AER} and a set of multi-purpose \ac{FPGA} routing boards.}{The implementation of event- or spike-based convolutional networks was first investigated in~\cite{Serrano-Gotarredona_etal99} with actual working chips implemented in the spike-based neuromorphic vision system ``CAVIAR''~\cite{Serrano-Gotarredona_etal09}. While the original chips in CAVIAR were designed using mixed-signal analog/digital circuits, the system is currently based on a custom digital chip implemented in a standard 0.35\,$\mu$m \ac{CMOS} process, containing an array of 32$\times$32 pixels with an arbitrary kernel size up to 32$\times$32~\cite{Camunas-Mesa_etal11}. Building large-scale networks requires an infrastructure for routing spikes between multiple convolutional chips.}  Currently convolutional networks are both  implemented on full custom \ac{ASIC} platforms~\cite{Conti_Benini15}, and on \ac{FPGA} platforms~\cite{Camunas-Mesa_etal12}.
The latest \ac{FPGA} implementation is done on a Xilinx Virtex-6, and supports up to 64 parallel convolutional modules of size 64$\times$64 pixels~\cite{Zamarreno-Ramos_etal13}. Here, memory is used to store the states of the pixels of the 64 parallel modules, and the 64 convolutional kernels of up to size 11$\times$11.  When a new event arrives, the kernel of each convolution module is projected around a pixel and a maximum of 121 updates are needed to determine the current state of the convolution. The new state is compared to a threshold to determine if an output spike should be generated, and its value is updated in memory. The event-driven convolutional modules have different memory requirements from the frame-based networks. Since the events arrive asynchronously, the states of the convolutional modules need to be stored in memory all the time. However, since events are processed in sequence, only a single computational adder block is needed, for computing the convolution of the  active pixel. A four-layered network which recognizes a small subset of digits was demonstrated using this implementation~\cite{Zamarreno-Ramos_etal13}. Scaling up the network further will require more logic gate resources or a custom digital platform which could support a much larger number of units, such as the one described in Section~\ref{sec:spinnaker}.


\subsection{Deep Belief Networks}
\label{sec:fpga-dbns}

\begin{figure}
\centering
\includegraphics[width=0.475\textwidth]{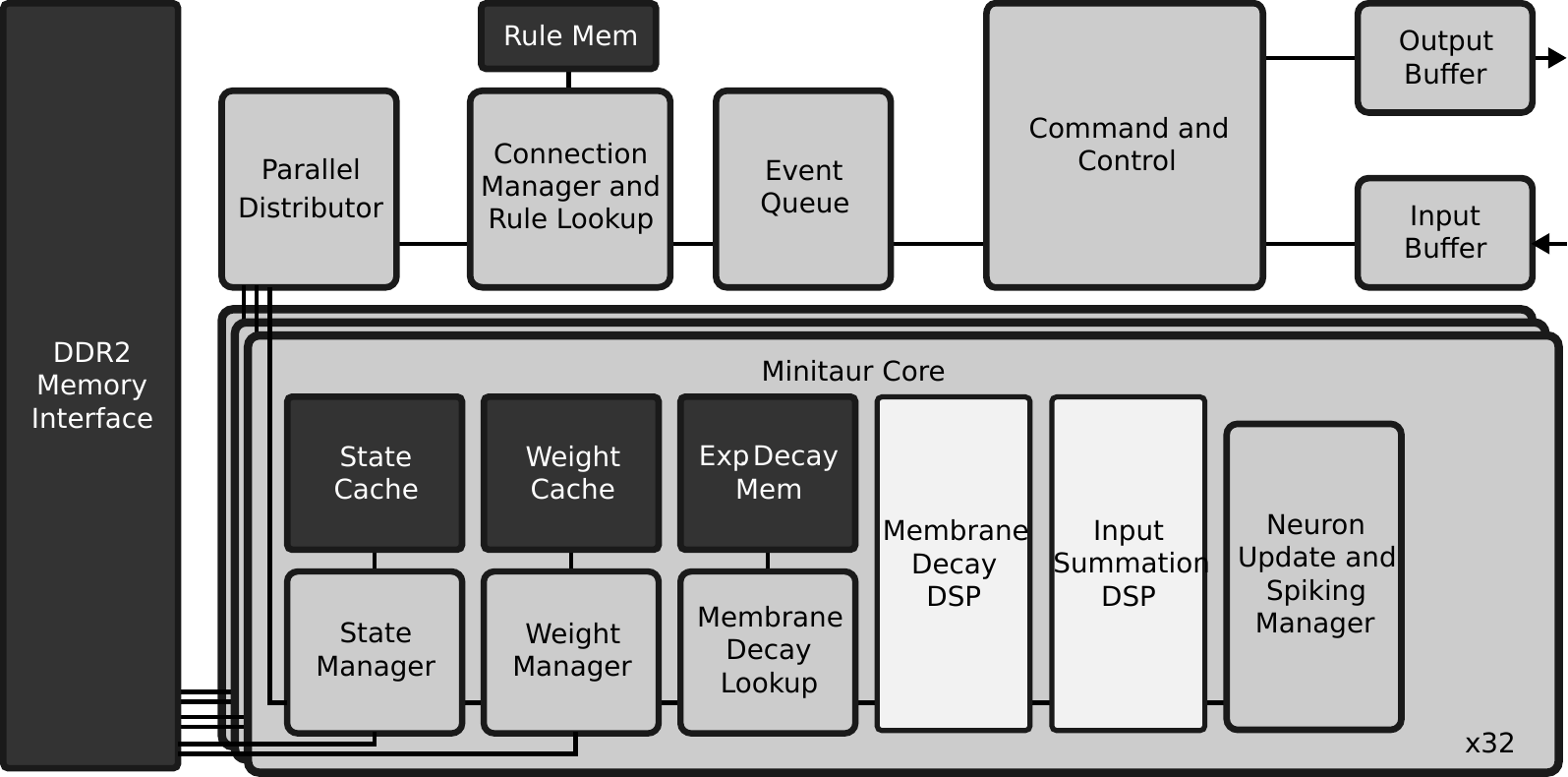}
\caption{Simplified architecture of the Minitaur system.  Each core has 2048\,B of state cache, 8192\,B of weight cache, and 2 \acp{DSP}: one for multiplying the decay, one for summation of the input current. The darker gray blocks indicate use of BRAMs. Events can be streamed from the computer via the input buffer and added to the event queue which also holds the events from the output neurons in the Minitaur Core. These events are then streamed out through the output buffer or to the neurons in the Core using the Connection Manager block.  Adapted from ~\cite{Neil_Liu14}.}
\label{fig:minitaurblock}
\end{figure}

\acp{DBN}, first introduced by Hinton and colleagues~\cite{Hinton_etal06} are a special class of deep neural networks with generative properties. They are composed of interconnected pairs  of Restricted Boltzmann Machines (RBMs, see Fig.~\ref{fig:multi-layer} (b)).  They have also been used in a variety of bench-marking tasks~\cite{Ciresan_etal10}. An adaptation of the neural model to allow transfer of parameters to a 784-500-500-10 layer spiking \ac{DBN} was described in~\cite{OConnor_etal13} with good performance on the MNIST\,digit database~\cite{mnist}. This network architecture has been implemented on a Xilinx Spartan-6 LX150 \ac{FPGA}~\cite{Neil_Liu14} with very similar classification performance results (92\%) on the same MNIST database.  
This \ac{FPGA} implementation of the \ac{DBN} (also called Minitaur) contains 32 parallel cores and 128\,MB of DDR2 as main memory (see Fig.~\ref{fig:minitaurblock}).  Each core has 2048\,B of state cache, 8192\,B of weight cache, and 2 \acp{DSP} for performing fixed-point math. Because of the typical all-to-all connection from neurons of one layer to the next projection layer, memory for storing the weights of these connections is critical. The cache locality of each of the 32 cores is critical to optimizing the neuron weights and the state lookups. The connection rule lookup block in Fig.~\ref{fig:minitaurblock} specifies how an incoming spike is projected to a set of neurons and the connection manager block distributes the updates of the neurons using the 32 parallel cores. This system is scalable but with limitations imposed by the number of available logic gates. On the Spartan-6 used the system can support up to 65,536 integrate-and-fire neurons.

The Minitaur system, as well as other deep network systems, operate by construction in a massively parallel way with each unit processing local data and using local memory resources. In order to efficiently map these architectures onto \acp{FPGA}, \acp{DSP}, or classical von Neumann systems, it is necessary to develop custom processing and memory optimized routing schemes that take into account these features~\cite{Farabet_etal11,Majumdar_etal12,Cassidy_etal13,Neil_Liu14}. 

\section{Large scale models of neural systems}
\label{sec:cust-memory-optim}

While dedicated implementations of convolutional and deep networks \revised{deepnn}{can be extremely}{are} useful for specific application domains such as visual processing and pattern recognition, they do not offer a computational substrate that is general enough to model information processing in complex biological neural systems. \revised{intro-iii}{To achieve this goal, it is necessary to develop spiking neural network architectures with some degree of flexibility, such as the possibility to configure the network connectivity, the network parameters, or even the models of the network's constituent elements (e.g., the neurons and synapses).}{}
\revised{general}{A common approach, that allows a high degree of flexibility, and that is}{A more general approach, that is nonetheless} closely related to the ones used for implementing convolutional and deep networks, is to implement generic spiking neural network architectures using  off-the-shelf \ac{FPGA} devices~\cite{Cassidy_etal13,Wang_etal13,Maguire_etal07}.  Such devices can be extremely useful for relatively rapid prototyping and testing of neural model characteristics because of \revised{flexibility}{their programmability}{the flexibility of these platforms in comparison to custom mixed-signal hybrid or complete digital VLSI solutions}. However, these devices, developed to implement conventional logic architectures with small numbers of input (fan-in) and output (fan-out) ports, do not allow designers to make dramatic changes to the system's memory structure, leaving the von Neumann bottleneck problem largely unsolved. \revised{asic}{The next level of complexity in the quest of implementing brain-like neural information processing systems is to design custom \acp{ASIC} using a standard digital design flow~\cite{Farahini_etal14}. Further customization can be done by combining standard design digital design flow for the processing elements, and custom asynchronous routing circuits for the communication infrastructure.}{}

\subsection{SpiNNaker}
\label{sec:spinnaker}

This is the current approach followed by the SpiNNaker\footnote{\revisednolines{SpiNNaker is a contrived acronym derived from Spiking Neural Network Architecture. The SpiNNaker project is lead by Prof. Steve Furber at Manchester University. It started in 2005 and was initially  funded by a UK government grant until early 2014. It is currently used as the ``many-core'' Neuromorphic Computing Platform for the EU FET Flagship Human Brain Project.}} project~\cite{Furber_etal14}. The SpiNNaker system is a multi-core computer designed with the goal of simulating the behavior of up to a billion neurons in real time. It is planned to integrate 57,600 custom VLSI chips, interfaced among each other via a dedicated global asynchronous communication infrastructure based on the \ac{AER} communication protocol~\cite{Mahowald94,Deiss_etal98,Boahen00} that supports large fan-in and large fan-out connections, and that has been optimized to carry very large numbers of small packets (e.g. representing neuron spikes) in real-time. Each SpiNNaker chip is a ``System-in-Package'' device that contains a VLSI die integrating 18 fixed-point \ac{ARM} ARM968 cores together with the custom routing infrastructure circuits, and a 128\,MB \ac{DRAM} die. In addition to the off-chip \ac{DRAM}, the chip integrates the router memory, consisting of a 1024$\times$32 3-state \ac{CAM} and a 1024$\times$24\,bit \ac{RAM} module.  Furthermore, each \ac{ARM} core within the chip comprises 64\,KB of data memory and 32\,KB of instruction memory. 

SpiNNaker represents a remarkable platform for fast simulations of large-scale neural computational models. It can implement networks with arbitrary connectivity and a wide variety of neuron, synapse, and learning models (or other algorithms not necessarily related to neural networks). However, the more complex the models used, the fewer number of elements that can be simulated in the system. In addition this system is built using  standard von Neumann computing blocks, such as the \ac{ARM} cores in each chip. As a consequence it uses to a large extent the same memory hierarchies and structures found in conventional computers (as in Fig.~\ref{fig:memory-pc}), and does not provide a computing substrate that can solve the von Neumann bottleneck problem~\cite{Backus78}.

\subsection{TrueNorth}
\label{sec:truenorth}

The recent implementation of a full custom spiking neural network \ac{ASIC} by IBM named ``TrueNorth''\footnote{\revisednolines{The development of the TrueNorth IBM chip was funded by the US  ``SyNAPSE''DARPA program, starting in November 2008.}}  represents a radical departure from classical von Neumann architectures~\cite{Merolla_etal14a}. Although the electronic circuits of TrueNorth  use transistors as digital gates, they are fully asynchronous and communicate using event-driven methods. The overall architecture consists of 4096 cores of spiking neural networks integrated into a single \ac{CMOS} chip. Each core comprises 256 digital leaky integrate and fire neuron circuits, 256$\times$256 binary programmable synaptic connections, and asynchronous encoding, decoding and routing circuits. Synaptic events can be assigned one of three possible strengths (e.g., to model one type of inhibitory synapse and two excitatory ones with different weights), but they are instantaneous pulses with no temporal dynamics. The dynamics of the neurons is discretized into 1\,ms time steps set by a global 1\,kHz clock. 
Depending on the core's synaptic matrix, the source neuron can target from one up to 256 neurons of a destination core. These routing schemes are not as flexible as in the SpiNNaker system, but as opposed to SpiNNaker, this architecture distributes the system memory, consisting of the core synaptic matrix and the routing tables entries, across the whole network. The system is inherently parallel, distributed, modular, and (by construction) fault-tolerant.
The cost of this very high parallelism however is relative density inefficiency: the chip fabricated using an advanced 28\,nm \ac{CMOS} process, occupies an area of 4.3\,cm$^2$, and all unused synapses in a given application represent ``dark silicon'' \revised{darksilicon}{(silicon area occupied by unused circuits). Note that since also in biology space is precious real-estate, unused synapses are typically removed by a dynamic process (see structural plasticity in Section~\ref{sec:plasticity-biology}). In the TrueNorth chip}{Furthermore,} the synapses do not implement any plasticity mechanism, so they cannot perform on-line learning or form memories. As a consequence, the goal of co-localizing memory and computation to mitigate the von Neumann bottleneck problem is only partially solved. 

\subsection{NeuroGrid}
\label{sec:neurogrid}

Similar to SpiNNaker and TrueNorth, the goal of the NeuroGrid\footnote{\revisednolines{The NeuroGrid project was developed by the group of Prof. Kwabena Boahen in Stanford, and was  funded by the US NIH Pioneer Award granted to Boahen in 2006.}} system~\cite{Benjamin_etal14} is to implement large-scale neural models and to emulate their function in real-time. Unlike the previous two approaches NeuroGrid follows the original neuromorphic engineering vision~\cite{Mead90,Douglas_etal95a} and uses analog/digital mixed-signal \revised{neurogrid1}{sub-threshold}{} circuits to model continuous time neural processing elements. In particular, important synapse and neuron functions, such as exponentiation, thresholding, integration, and temporal dynamics are directly \emph{emulated} using the physics of \acp{FET} biased in the sub-threshold regime~\cite{Liu_etal02a}.

\begin{figure}
\centering
\includegraphics[width=0.4\textwidth]{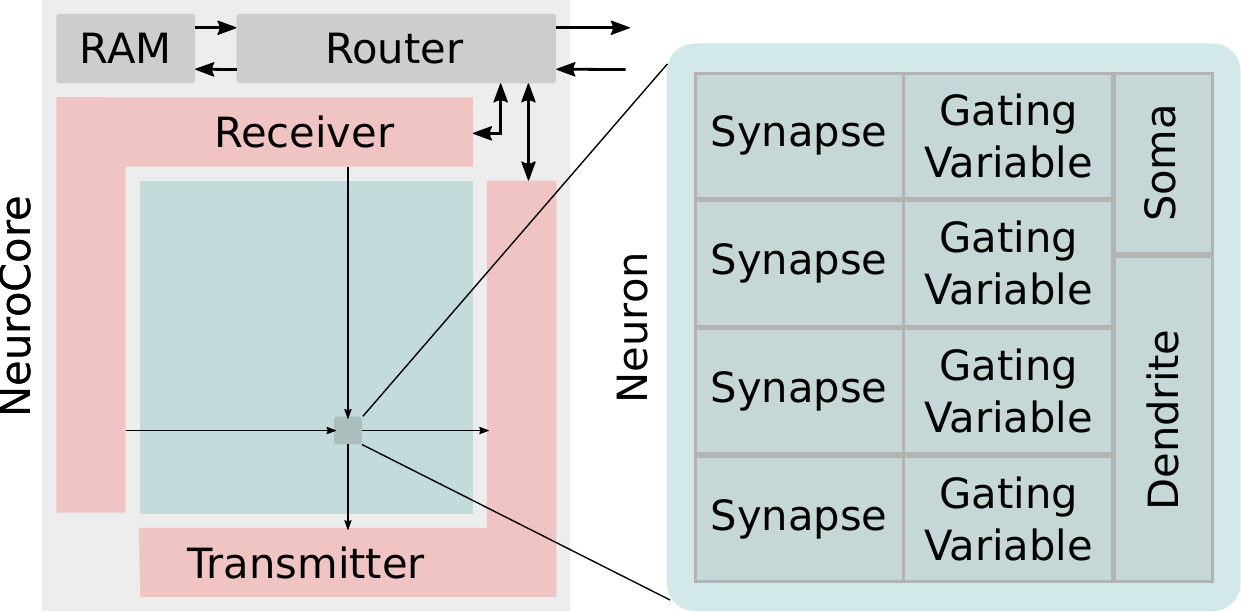}
\caption{NeuroCore chip block diagram (adapted from~\cite{Benjamin_etal14}). The chip comprises a 256$\times$256 array of neuron elements, an asynchronous digital transmitter for sending the events generated by the neurons, a  receiver block for accepting events from other sources, a router block for communicating packets among chips, and a memory blocks for supporting different network configurations. The neuron block comprises four different types of synapse analog circuits that integrate the incoming digital events into analog currents over time, four analog gating variable circuits that model the ion channel population dynamics, a soma circuit that generates the neuron output spikes, and a dendritic circuit that integrates the synaptic currents over space, from neighboring neurons.}
\label{fig:neurogrid}
\end{figure}

NeuroGrid consists of a board with 16 standard \ac{CMOS} ``NeuroCore'' chips connected in a tree network, with each NeuroCore consisting of a 256$\times$256 array of two-compartmental neurons \revised{neurocore}{(see Fig.~\ref{fig:neurogrid})}{}. Synapses are ``shared'' among the neurons by using the same synapse circuit for different spike sources. Multiple spikes can be superimposed in time onto a single synapse circuit, because it has been designed as a linear integrator filter, and no non-linear effects are modeled. Each neuron in the array can target multiple destinations thanks to an asynchronous multi-cast tree routing digital infrastructure. The number of target destinations that a neuron can reach is limited by the size of the memory used in external routing tables and by its access time~\cite{Choudhary_etal12}. However NeuroGrid increases the fan-out of each neuron by connecting neighboring neurons with local resistive networks or diffusors that model synaptic gap-junctions~\cite{Boahen_Andreou92}.  This structured synaptic organization is modeled after the layered organization of neurons within cortical columns~\cite{Binzegger_etal04}. The full NeuroGrid board therefore can implement models of cortical networks of up to one million neurons and billions of synaptic connections with sparse long range connections and dense local connectivity profiles.  Like TrueNorth, NeuroGrid represents a radical departure from the classical von Neumann computing paradigm. Different memory structures are distributed across the network (e.g., in the form of routing tables, parameters, and state variables). The ability of the shared synapses to integrate incoming spikes reproducing biologically plausible dynamics provide the system with computational primitives that can hold and represent the system state for tens to hundreds of milliseconds. However, the design choice to use linear synapses in the system excluded the possibility to implement synaptic plasticity mechanisms at each synapse, and therefore the ability of NeuroGrid to model on-line learning or adaptive algorithms without the aid of additional external computing resources.

NeuroGrid has been designed to implement cortical models of computation that run in real-time, and has been used in a closed-loop brain-machine application~\cite{Dethier_etal11} and to control articulated robotic agents~\cite{Menon_etal14}. In this system \revised{virtualize}{time represents itself~\cite{Mead89}, and data is processed on the fly, as it arrives. Computation is data driven and signals are consumed as they are received. Unlike in conventional von Neumann architectures, time is not ``virtualized'': signals are not time-stamped and there are no means to store the current state of processing or to transfer time-stamped partial results of signal processing operations to external memory banks for later consumption. Memory and computation are expressed in the dynamics of the circuits, and in the way they are interconnected.}{it is not possible to virtualize time (here ``time represents itself''~\cite{Mead89})} So it is important that the system's memory and computing resources have time-constants that are well matched to those of the signals they  process\revised{virtualize2}{}{ in real-time}. As the goal  is to interact with the environment and process natural signals with biological time-scales, these circuits use biologically realistic time constants which are extremely long (e.g., tens of milliseconds) if compared to the ones used in typical digital circuits. This long time-constants constraint is not easy to achieve using conventional analog \ac{VLSI} design techniques. Achieving this goal, while minimizing the size of the circuits (to maximize density), is possible only if one uses extremely small currents, such as those produced by transistors biased in the sub-threshold domain~\cite{Liu_etal02a,Chicca_etal14}, \revised{neurogrid2}{as it is done in NeuroGrid}{}.

\subsection{BrainScales}
\label{sec:brainscales}

Another approach for simulating large-scale neural models is the one being pursued in the BrainScales\footnote{\revisednolines{The BrainScales acronym stands for ``Brain-inspired multiscale computation in neuromorphic hybrid systems''. This project was funded by the European Union under the FP7-ICT program, and started in January 2011. It builds  on the research carried out in the previous EU FACETS (Fast Analog Computing with Emergent Transient States) project, and is now part of the EU FET Flagship Human Brain Project (HBP). The mixed signal analog-digital system being developed in this project  currently represents the ``physical model'' Neuromorphic Computing Platform of the HBP.}} project~\cite{BrainScales}.  BrainScales aims to implement a  wafer-scale neural simulation platform, in which each 8\,inch silicon wafer integrates 
50$\times$106 plastic synapses and 200,000 biologically realistic neuron circuits. The goal of this project is to build a custom mixed signal analog/digital simulation engine that can accurately implement the differential equations of the computational neuroscience models provided by  neuro-scientists, and reproduce the results obtained from numerical simulations run on standard computers as faithfully as possible. For this reason, in an attempt to improve the precision of the analog circuits, the BrainScales engineers chose to use the above-threshold, or strong-inversion, regime for implementing models of neurons and synapses. However, in order to  maximize the number of processing elements in the wafer, they chose to implement relatively small capacitors for  modeling the synapse and neuron capacitances. As a consequence, given the large currents produced by the above-threshold circuit and the small capacitors, the BrainScales circuits cannot achieve the long time-constants required for interacting with the environment in real-time. Rather, their dynamics are ``accelerated'' with respect to typical biological times by a factor of $10^3$ or $10^4$. This has the advantage of allowing very fast simulation times which can be useful e.g., to investigate the evolution of network dynamics over long periods of time, once all the simulation and network configuration parameters have been uploaded to the system. But it has the disadvantage of requiring very large bandwidths and fast digital, high-power, circuits for transmitting and routing the spikes across the network. 

Like NeuroGrid,  the synaptic  circuits in BrainScales express temporal dynamics, so they form memory elements that  can store the state of the network (even if for few hundreds of micro-seconds). In addition, the BrainScales synapses comprise also circuits endowed with spike-based plasticity mechanisms that allow the network to learn and form memories~\cite{Schemmel_etal07}. BrainScales therefore implements many of the principles that are needed to build brain inspired information processing systems that can replace or complement conventional von Neumann computing systems. But given the circuit design choices made for maximizing precision in reproducing numerical simulation results of given differential equations,  the system is neither low power (e.g., when compared to the other large-scale neural processing systems previously described), nor compact. To build  neural information processing systems that are at the same time compact, low-power, and robust, it will be necessary to follow an approach that can use extremely compact devices (such as nano-scale memristive synapse elements), very low-power circuit design approaches (e.g., with sub-threshold current-mode designs), and by adopting adaptation and learning techniques that can compensate for the variability and in-homogeneity present in the circuits at the system level.

\begin{table*}
  \centering
  \begin{tabular}{@{}p{0.125\textwidth} m{0.25\textwidth} m{0.45\textwidth}@{}}
    \toprule
    \textbf{Relative time scale}&\textbf{Computers}&\textbf{Brains and Neuromorphic Systems}\\
    \midrule
    Short & Fast local memory (e.g.,  registers and cache memory banks inside the \ac{CPU}) & Synapse and neural dynamics at the single neuron level (short-term plasticity, spike-frequency adaptation, leak, etc.)\\
    Medium & Main memory (e.g., dynamic RAM) & Spike-based plasticity (\acs{STDP}, Hebbian learning, etc.) Formation of  working memory circuits (e.g., recurrent networks, attractors, etc.)\\
    Long &  Peripheral memory (e.g., hard-disks) & Structural plasticity, axonal growth, long-term changes in neural pathways.\\
    \bottomrule
  \end{tabular}
  \caption{Memory structures in computers versus memory structures in brains and neuromorphic systems.}
  \label{tab:memory}
\end{table*}

\section{Adaptation, learning, and working-memory}
\label{sec:learning}

Adaptation and learning mechanisms in neural systems are mediated by multiple forms of ``plasticity'', which operate on a wide range of  time-scales~\cite{Feldman09}. The most common forms are Structural Plasticity, Homeostatic Plasticity, \ac{LTP} and \ac{LTD} mechanisms, and \ac{STP} mechanisms.
\revised{wm}{While these mechanisms are related to the ability of single neurons and synapses to form memories, the term \emph{Working Memory} is often used to refer to the ability of full networks of neurons to temporarily store and manipulate information. 
A common model that has been proposed to explain the neural basis of working memory is that based on ``Attractor Networks''~\cite{Hopfield82,Amit92,Renart_etal03a,Rolls10}. 
In this section we give a brief overview of single neuron and network level mechanisms observed in biology that sub-serve the function of memory formation, and of the neuromorphic engineering approaches that have been followed to implement them in electronic circuits. Examples of analogies between the memory structures in conventional von Neumann architectures and the plasticity and memory mechanisms found in neural and neuromorphic systems are shown in Table~\ref{tab:memory}.
}{}

\subsection{Plasticity}
\label{sec:plasticity-biology}

\emph{Structural plasticity} refers to the brain's ability to make physical changes in its structure as a result of learning and experience~\cite{Lamprecht_LeDoux04,Maguire_etal00}. This mechanism, which typically operates on very long time scales, ranging from minutes to days or more, is important for the formation and maintenance of long-term memories. \emph{Homeostatic plasticity} is a self-stabilizing mechanism that is used to  keep the activity of the neurons within proper operating bounds~\cite{Turrigiano_Nelson04}. It is a process that typically operates on relatively long time scales, ranging from hundreds of milliseconds to hours. From the computational point of view, this mechanism plays a crucial role for adapting to the overall activity of the network,  while controlling its stability. \emph{Short-term plasticity} on the other hand is a process that typically operates on time scales that range from fractions of milliseconds to hundreds of milliseconds. It can manifest itself as both short-term facilitation or short-term depression, whereby the strength of a synapse connecting a pre-synaptic (source) neuron to a post-synaptic (destination) one is up-regulated or down-regulated respectively, with each spike~\cite{Zucker_Regehr02}. It has been demonstrated that this mechanism can play a fundamental role in neural computation~\cite{Renart_etal03a,Buonomano00,Abbott_etal97}, e.g., for modulating the neuron's sensitivity to its input signals. Finally, \emph{long-term plasticity} is the mechanism responsible for producing long-lasting, activity dependent changes in the synaptic strength of individual synapses~\cite{Hebb49}. 
A popular class of \ac{LTP} mechanisms that has been the subject of widespread interest within the neuroscience community~\cite{Markram_etal12}, the neuromorphic community~\cite{Azghadi_etal14}, and more recently in the material science and nano-technology community~\cite{Serrano-Gotarredona_etal13,Saighi_etal15}, is based on the \ac{STDP} rule~\cite{Markram_etal97,Abbott_Nelson00}.  In its simplest form, the relative timing between the pre- and post-synaptic spikes determines how to update the efficacy of a synapse. In more elaborate ones, other factors are taken into account, such as the average firing rate of the neurons~\cite{Gjorgjieva_etal11}, their analog state (e.g., the neuron's membrane potential)~\cite{Fusi02,Graupner_Brunel12}, and/or the current value of the synaptic weights~\cite{Sjostrom_etal01,Nessler_etal13}. The time-scales involved in the \ac{STDP} timing window for the single weight update are of the order of tens to hundreds of milliseconds. But the \ac{LTP} and \ac{LTD} changes induced in the synaptic weights last for much longer time scales, ranging from hours to weeks~\cite{Fusi_etal05}. 

\subsection{Attractor networks}
\label{sec:attractor-networks}

Mechanisms operating at the network level can also allow neural processing systems to form short-term memories, consolidate long-term ones, and carry out non-linear processing functions such as selective amplification (e.g., to implement attention and decision making). 
An example of such a network-level mechanism is provided by ``attractor networks''. 
These are networks of neurons that are recurrently connected via excitatory synapses, and that can settle into stable patterns of firing even after the external stimulus is removed. Different stimuli can elicit different stable patterns, which consist of specific subsets of neurons firing at high rates. Each of the high-firing rate attractor states can represent a different memory~\cite{Amit92}. To make an analogy with conventional logic structures, a small attractor network with two stable states would be equivalent to a flip-flop gate in \ac{CMOS}. 
\begin{figure}
  \centering
  \includegraphics[width=0.45\textwidth]{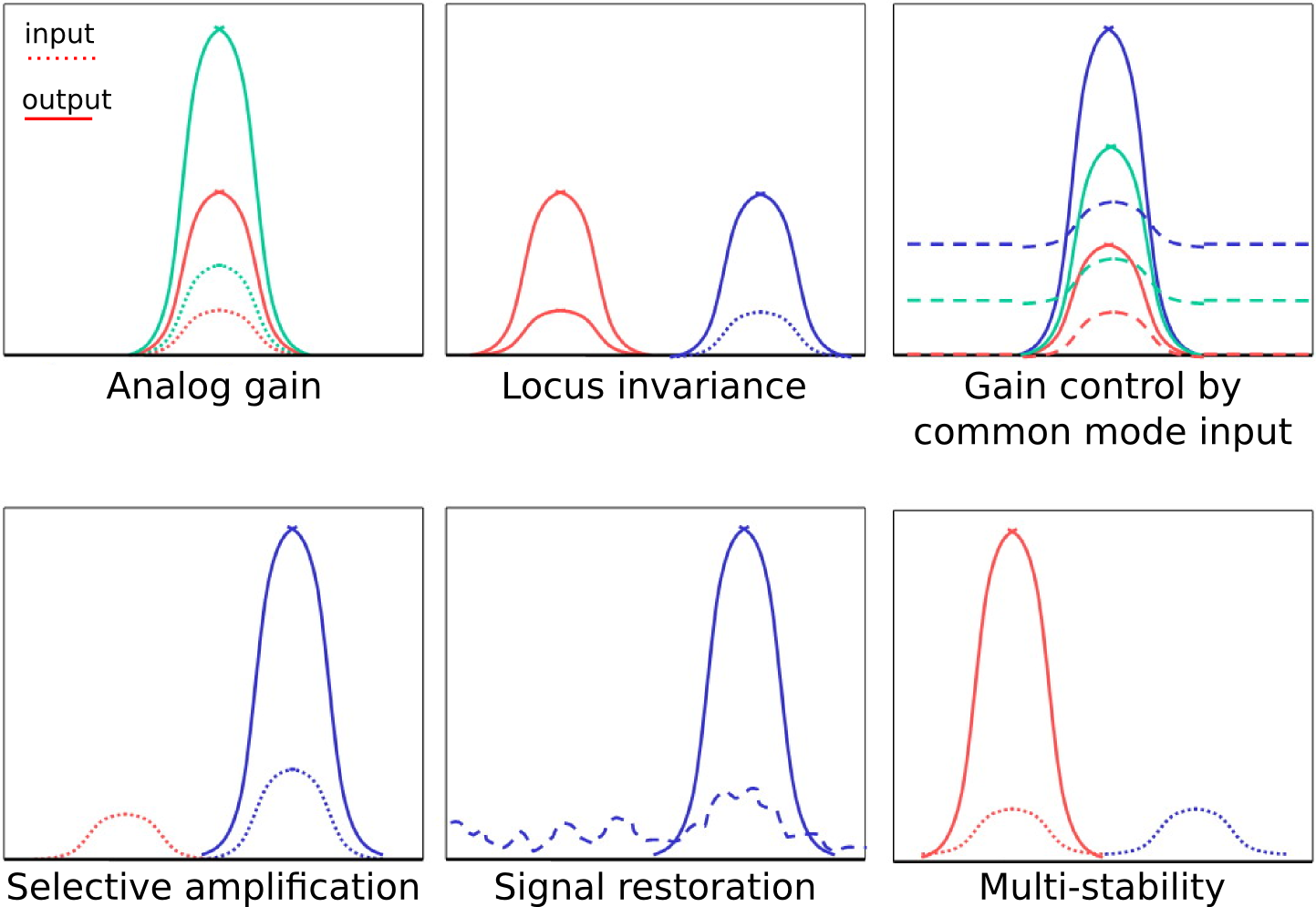}
  \caption{Soft Winner-Take-All network behaviors: linear (top row) and non-linear (bottom row). The horizontal axis of each trace represents the spatial location of the neurons in the network, while the vertical axis represents the neurons response amplitude. Figure adapted from~\cite{Indiveri_etal09}.}
  \label{fig:wta}
\end{figure}

A particularly interesting class of attractor networks is the one of  \ac{sWTA} neural networks. In these networks, groups of neurons both cooperate and compete with each other. Cooperation takes place between groups of neurons spatially close to each other, while competition is typically achieved through global recurrent patterns of inhibitory connections~\cite{Yuille_Geiger03}. When stimulated by external inputs, the neurons excite their neighbors and the ones with highest response suppress all other neurons to win the competition. Thanks to these competition and cooperation mechanisms, the outputs of individual neurons depend on the activity of the whole network and not just on their individual inputs~\cite{Douglas_etal95}. Depending on their parameters and input signals, \acp{sWTA} networks can perform both linear and complex non-linear operations~\cite{Douglas_Martin07} (see Fig.~\ref{fig:wta}), and have been shown to posses powerful computational properties for tasks involving feature-extraction, signal restoration and pattern classification~\cite{Maass00}.
Given their structure and properties, they have been proposed as canonical microcircuits that can explain both the neuro-anatomy and the neuro-physiology data obtained from experiments in the mammalian cortex~\cite{Douglas_Martin04}. \revised{dnf}{Such networks have also been linked to the \acp{DNF} typically used to model behavior and cognition in autonomous agents~\cite{Sandamirskaya13,Schoner07}}{}.

\subsection{Neuromorphic circuit implementations}
\label{sec:plast-mech-neur}

The neuromorphic engineering community has been building physical models of \ac{sWTA} networks~\cite{Indiveri_etal01,Chicca06,Oster_etal09}, attractor networks~\cite{Chicca_etal03,Giulioni_etal12}, and plasticity mechanisms~\cite{Azghadi_etal14} that cover the full range of temporal and spatial scales described in Section~\ref{sec:plasticity-biology} for many years. For example, several circuit solutions have been proposed to implement short-term plasticity dynamics, using different types of devices and following a wide range of design techniques~\cite{Rasche_Hahnloser01, Boegerhausen_etal03, Bill_etal10,  Noack_etal11, Ohno_etal11, Dowrick_etal12}; a large set of spike-based learning circuits have been proposed to model long-term plasticity~\cite{Indiveri03b, Bofill-i-Petit_Murray04, Riis_Hafliger04, Indiveri_etal06, Schemmel_etal07, Arthur_Boahen08, Mitra_etal09, Giulioni_etal09, Bamford_etal12, Ramakrishnan_etal12,Azghadi_etal14}; multiple solutions have been proposed for implementing homeostatic plasticity mechanisms~\cite{Bartolozzi_Indiveri09,Rovere_etal14}; impressive demonstrations have been made showing the properties of \ac{VLSI} attractor networks~\cite{Chicca_etal03,Giulioni_etal12,Pfeil_etal13,Chicca_etal14}; while structural plasticity has been implemented both at the single chip level,
with morphology learning mechanisms for dendritic trees~\cite{Hussain_etal14} and at the system level, in multi-chip systems that transmit spikes using the \ac{AER} protocol, by reprogramming or ``evolving'' the network connectivity routing tables stored in the digital communication infrastructure memory banks~\cite{Fasnacht_Indiveri11,Chicca_etal07a}. While some of these principles and circuits have been adopted in the deep network implementations of Section~\ref{sec:electronic-systems} and in the large-scale neural network implementations of Section~\ref{sec:cust-memory-optim}, many of them still remain to be exploited, at the system and application level, for endowing neuromorphic systems with additional powerful computational primitives.

\section{A neuromorphic processor}
\label{sec:neur-proc}

\begin{figure}
  \centering
  \includegraphics[width=0.45\textwidth]{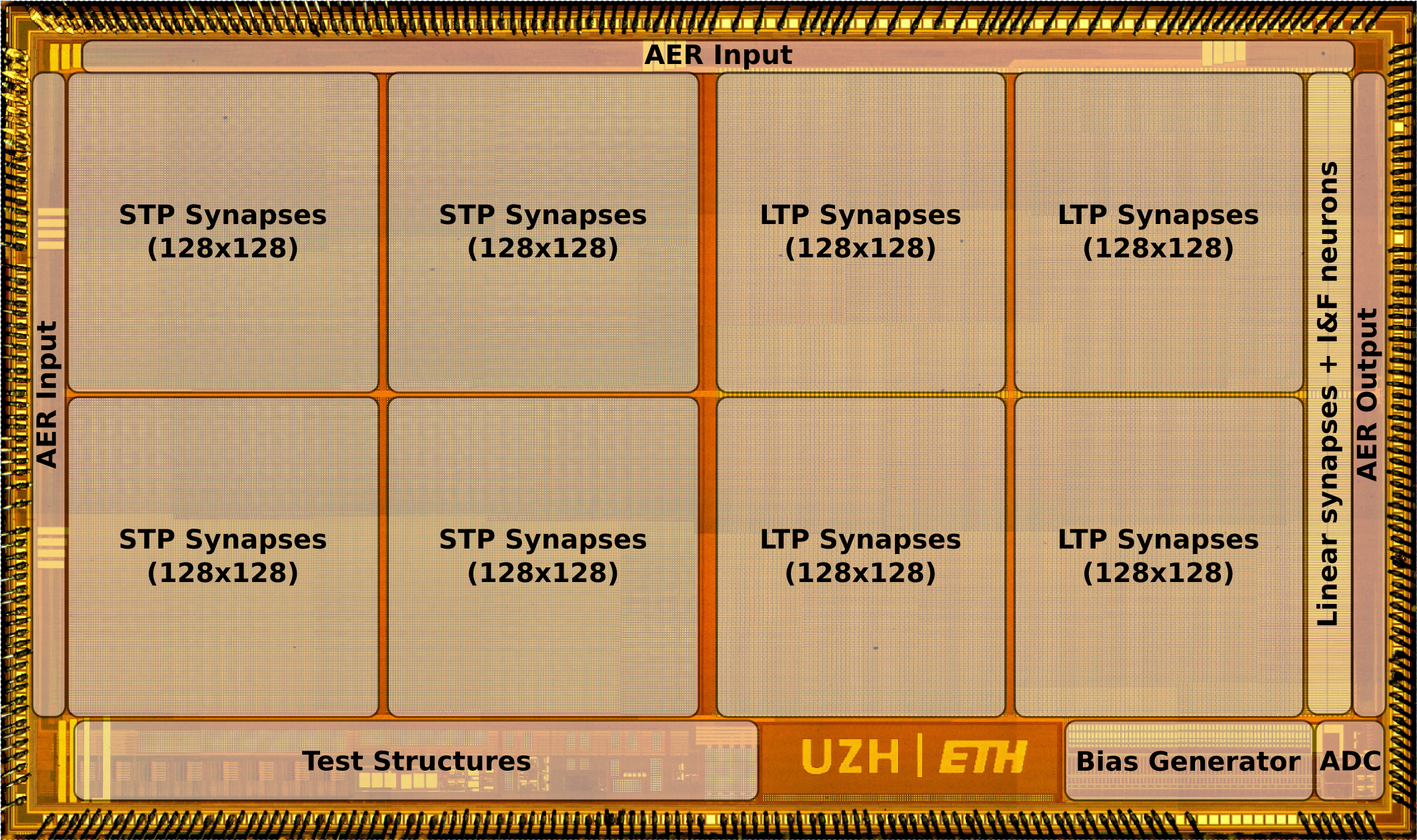}
  \caption{\ac{NP}: micrograph of a neuromorphic processor chip that allocates most of its area to non-linear synapse circuits for memory storage and distributed massively parallel computing.}
  \label{fig:rolls}
\end{figure}

An example of a recently proposed neuromorphic multi-neuron chip that integrates all of the mechanisms described in Section~\ref{sec:learning} is the \ac{NP}\footnote{\revisednolines{The \ac{NP} device was designed at the Institute of Neuroinformatics of the University of Zurich, ETH Zurich. Its development was funded by the EU ERC ``Neuromorphic Processors'' (NeuroP) project, awarded to Giacomo Indiveri in 2011.}}~\cite{Qiao_etal15}. This device implements a configurable spiking neural network using slow sub-threshold neuromorphic circuits that directly emulate the physics or real neurons and synapses, and fast asynchronous digital logic circuits that manage the event-based \ac{AER} communication aspects as well as the properties of neural network. While the analog circuits faithfully reproduce the neural dynamics and the adaptive and learning properties of neural systems, the asynchronous digital circuits provide a flexible means to configure both parameters of the individual synapse and neuron elements in the chip, as well as the connectivity of the full network. The goal of the approach followed in designing this device was not to implement large numbers of neurons or large-scale neural networks, but to integrate many non-linear synapses for exploring their distributed memory and information processing capabilities. \revised{essentially1}{}{The device is essentially a large memory chip in which the memory elements are complex synapse circuits.} \revised{mismatch}{Although the analog circuits in the \ac{NP} are characterized by susceptibility to noise, variability and inhomogeneous properties (mainly due to device mismatch)~\cite{Qiao_etal15}}{The} the multiple types of plasticity mechanisms and the range of temporal dynamics present in these circuits endow the system with a set of collective and distributed computational operators that allow it to implement a wide range of \revised{mismatch2}{robust}{} signal processing and computing functions~\cite{Qiao_etal15}. The device block diagram and chip micro-graph is depicted in Fig.~\ref{fig:rolls}. \revised{essentially2}{As evidenced from the chip micrograph in Fig.~\ref{fig:rolls}, most of the area of the device is dedicated to the the synapses, which represent both the site of memory and of computation.}{}

The chip, fabricated using a standard 6-metal 180\,nm \ac{CMOS} process, occupies an area of 51.4\,mm$^2$ and has approximately 12.2 million transistors. It comprises 256 neurons and 133,120 synapses, equivalent to 130\,KB ``memory'' elements. The synapse circuits are of three different types: linear time-multiplexed (shared) synapses, \ac{STP} synapses, and \ac{LTP} synapses. The linear synapses are subdivided into blocks of four excitatory and four inhibitory synapse integrator circuits per neuron, with shared sets of synaptic weights and time constants. The \ac{STP} synapses are arranged in four arrays of 128$\times$128 elements. Each of these elements has both analog circuits, that can reproduce short-term adaptation dynamics, and digital circuits, that can set and change the programmable weights. The \ac{LTP} synapses are subdivided into four arrays of 128$\times$128 elements which contain \revised{analog-digital}{both analog learning circuits, and digital state-holding logic. The learning circuits implement the  stochastic plasticity \ac{STDP} model proposed in~\cite{Brader_etal07} to update the synaptic weight upon the arrival of every pre-synaptic input spike. Depending on the analog value of the weight, the learning circuits also drive the weight to either a high \ac{LTP} state, or a low \ac{LTD} state on very long time scales (i.e., hundreds of milliseconds), for long-term stability and storage of the weights (see~\cite{Qiao_etal15} for a through description and characterization of these circuits). The digital logic in the \ac{LTP} synapse elements is used for configuring the network connectivity.}{}
The silicon neuron circuits \revised{neurons}{on the right side of the layout of Fig.~\ref{fig:rolls}}{} implement a model of the adaptive exponential \ac{IF} neuron~\cite{Livi_Indiveri09} that has been shown to be able to accurately reproduce electrophysiological recordings of real neurons~\cite{Rossant_etal10,Brette_Gerstner05}. Additional circuits are included in the \ac{IF} neuron section of Fig.~\ref{fig:rolls} to drive the learning signals in the \ac{LTP} arrays, and to implement self-tuning synaptic scaling homeostatic mechanism~\cite{Turrigiano08} on very long time scales (i.e., seconds to minutes)~\cite{Rovere_etal14}. \revised{all-synapses}{The currents produced by the synapse circuits in the \ac{STP} and \ac{LTP} arrays}{All synaptic currents} are integrated by \revised{all-synapses2}{two independent sets of}{} low-power log-domain pulse integrator filters~\cite{Bartolozzi_Indiveri07a} that can reproduce synaptic dynamics with time constants that can range from fractions of micro-seconds to hundreds of milliseconds. \revised{no-demux}{}{A static logic de-multiplexer circuit allows users to allocate the \ac{STP} and \ac{LTP} synaptic resources to neurons (e.g., from the default case of 512 synapses per neuron with all neurons operative, to 1024 synapses per neuron with only half of the neurons operative, all the way to 128\,KB per neuron, with only that neuron operative).} 
The programmable digital latches in the synapse elements can be used to set the state of the available all-to-all network connections, therefore allowing the user to configure the system to implement arbitrary network topologies, with the available 256 on-chip neurons, ranging from multi-layer deep networks, to recurrently connected reservoirs, to winner-take-all networks, etc.  All analog parameters of synapses and neurons can be configured via a temperature compensated programmable bias generator~\cite{Delbruck_etal10}.

\begin{figure*}
  \begin{subfigure}{\textwidth}
    \centering
    \includegraphics[width=\textwidth]{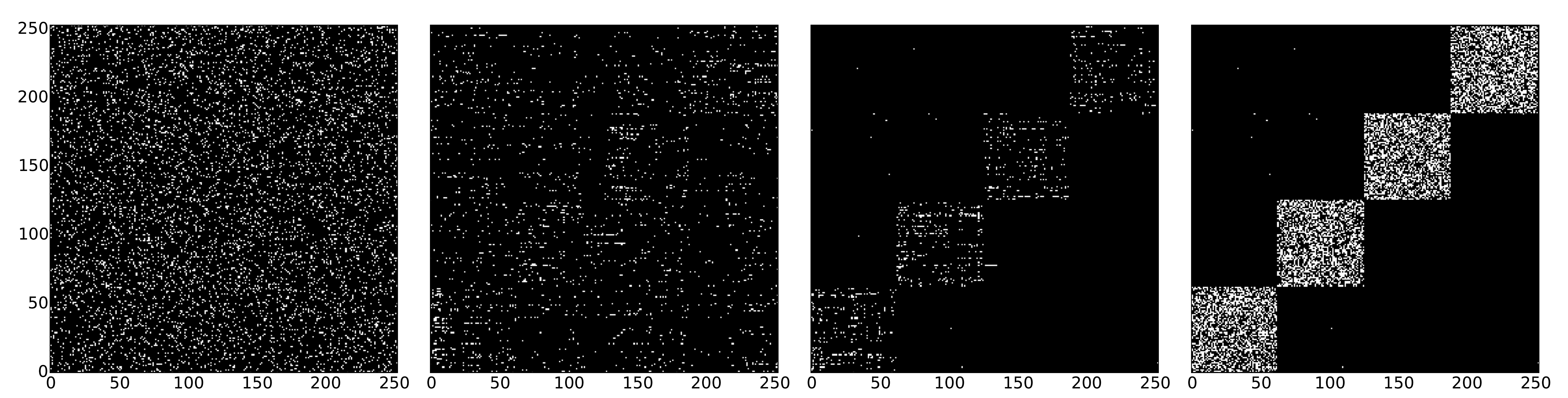}
    \subcaption{} 
    \label{fig:attractor-mat}
  \end{subfigure}\\
  \begin{subfigure}{\textwidth}
    \centering
    \includegraphics[width=\textwidth]{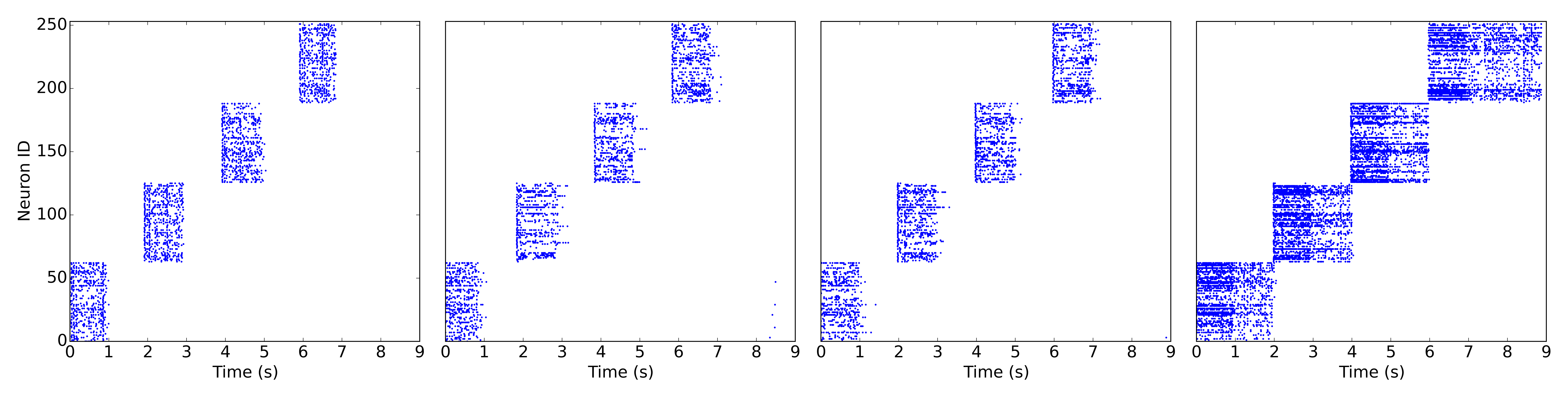}
    \subcaption{} 
    \label{fig:attractor-rasters}
  \end{subfigure}\\
  \begin{subfigure}{\textwidth}
    \centering
    \includegraphics[width=0.328\textwidth]{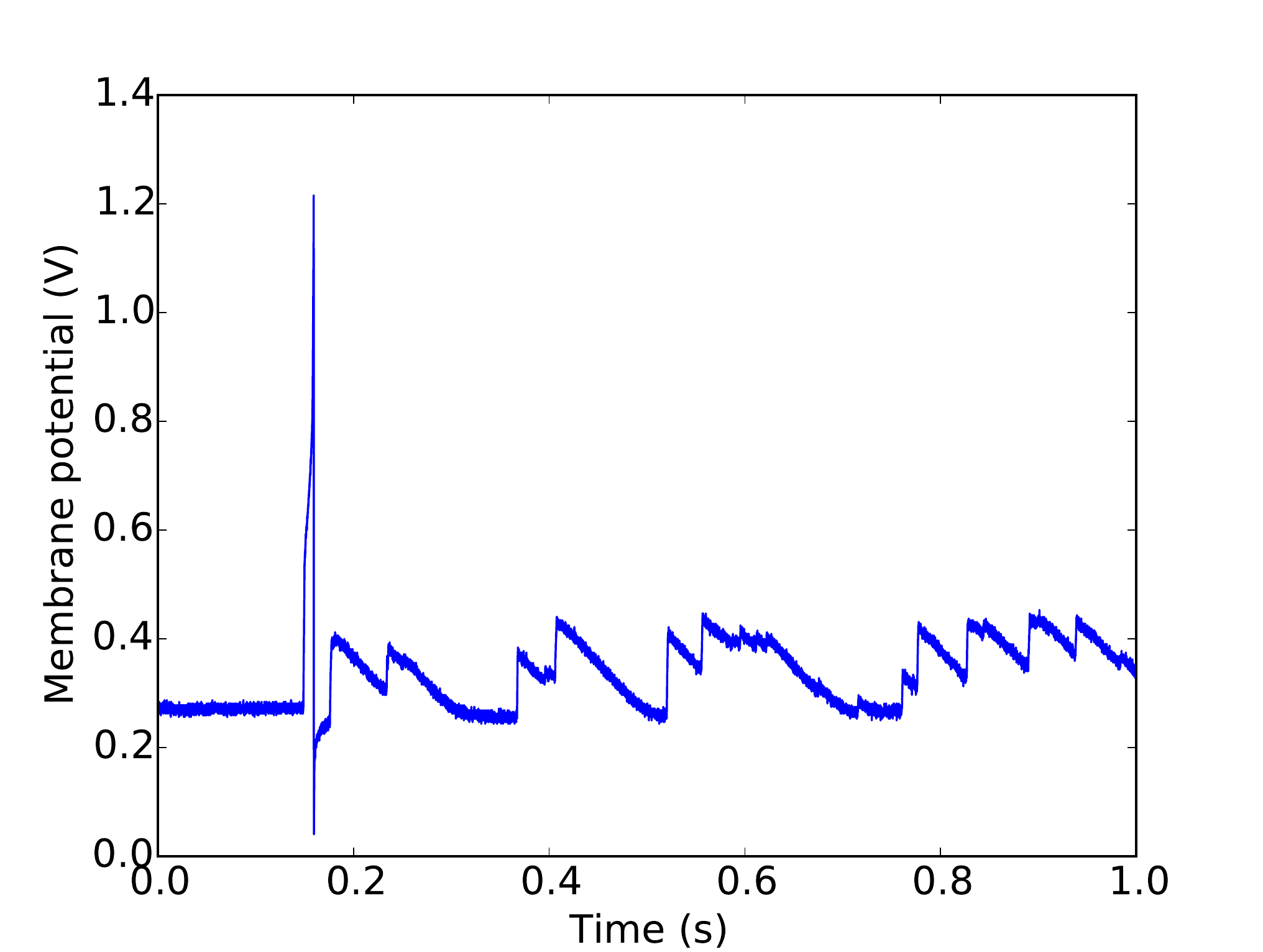}
    \includegraphics[width=0.328\textwidth]{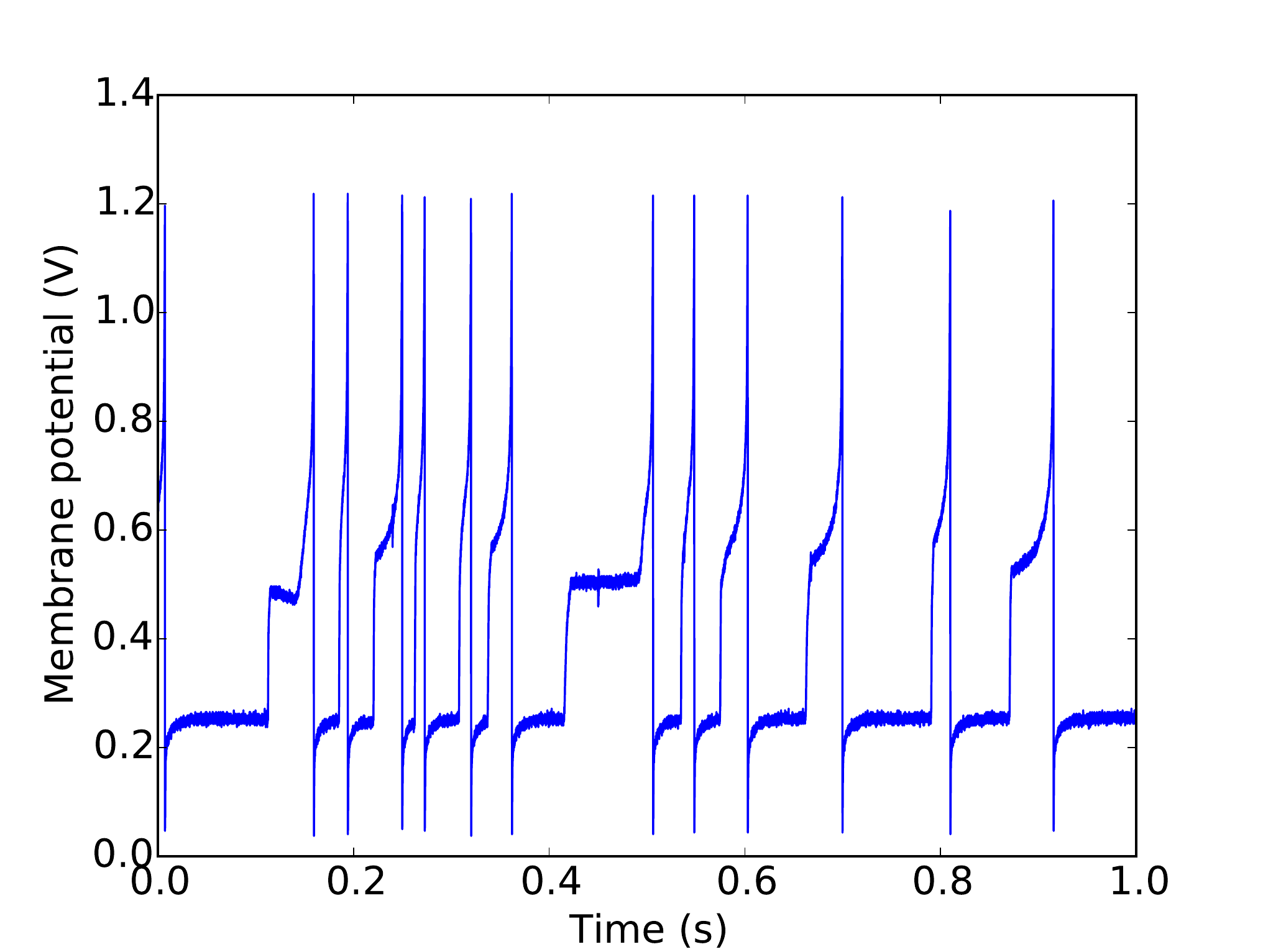}
    \includegraphics[width=0.328\textwidth]{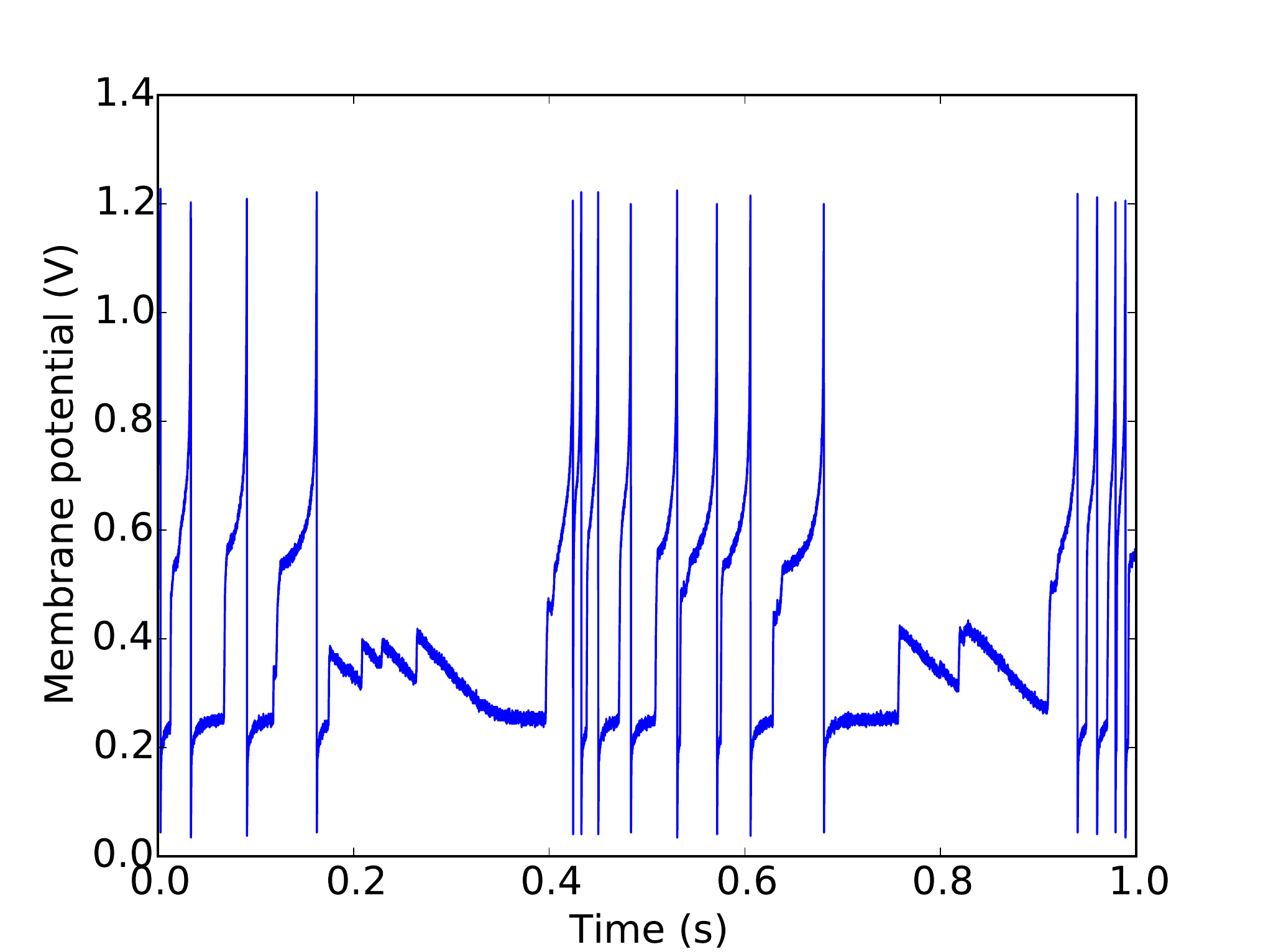}
    \subcaption{} 
    \label{fig:attractor-vmem}
  \end{subfigure}
  \caption{Forming stable attractors in the \ac{NP} (\subref{fig:attractor-mat}) State of the bi-stable \ac{LTP} synapses at the beginning, during, and at the end of the experiment; (\subref{fig:attractor-rasters}) Raster plots representing the response of the neurons, both during input stimulus presentation (for one second at t=0\,s, 2\,s, 4\,s, 6\,s), and after stimulus is removed; (\subref{fig:attractor-vmem}) Example of a measured neuron membrane potential at the beginning, during, and at the end of the training sessions, during stimulus presentation.}
  \label{fig:attractors}
\end{figure*}

\revised{demonstration}{
To demonstrate the neuromorphic processor's memory and information processing abilities we trained the network to encode memory patterns into four different attractor networks, as those described in Section~\ref{sec:attractor-networks}  (see Fig.~\ref{fig:attractors}). The protocol followed to train the silicon neurons to form these associative memories is the following: we created a recurrent competitive network, by connecting  the chip's 256 neuron outputs to 90\% of their 256$\times$256 plastic \ac{LTP} synapses, via recurrent excitatory connections, and to a random subset of 50\% of the non-plastic \ac{STP} inhibitory synapses, via recurrent inhibitory connections; we initialized the \ac{LTP}  bi-stable plastic synapses to a low state with 90\% probability, and to a high state with 10\% probability (see first plot of Fig.~\ref{fig:attractor-mat});  we configured the parameters of the learning circuits to induce \ac{LTP} and \ac{LTD} transitions in a intermediate range of firing rates (e.g., between 50\,Hz and 150\,Hz), and to stop changing the weights for higher frequencies (see~\cite{Qiao_etal15} for a detailed description of the algorithm and circuits that implement these features); we then stimulated four separate groups of neurons repeatedly, by stimulating the non-plastic synapses with Poisson distributed input spike trains, for one second each (see Fig.~\ref{fig:attractor-rasters}). With each stimulus presentation, the plastic synapses of the neurons receiving both feed-forward input from the Poisson input spike trains and recurrent feed-back from the output neurons tended to potentiate, while the plastic synapses of the neurons that did not receive feed-back spikes correlated with feed-forward inputs tended to depress (see second, third and fourth plot of Fig.~\ref{fig:attractor-mat}). As the number of potentiated plastic synapses increased, the populations of recurrently connected neurons started to produce sustained activity, with higher firing rates and more structured response properties (see second, third and fourth plot of Fig.~\ref{fig:attractor-mat}). The attractors are fully formed when enough recurrent connections potentiate, such that the firing rate of the stimulated population is high enough to stop the learning process and the population activity remains sustained even after the input stimulus is removed (e.g. see activity during t=1-2s\,s, t=3-4\,s, t=5-6\,s, and t=7-9\,s in Fig.~\ref{fig:attractor-rasters}). When neurons belonging to different attractors are stimulated, they suppress activity of all other attractors via the recurrent inhibitory connections. Figure~\ref{fig:attractor-vmem} shows an example of a silicon neuron output trace measured at the beginning of the experiment, when no attractors exist, in the middle, as the attractors are being formed, and at the end of the experiment, during sustained activity of a fully formed attractor. Note that although the analog circuits in the device have a high degree of variability (with a coefficient of variation of about 10\%~\cite{Qiao_etal15}), and that the learning process is stochastic~\cite{Brader_etal07}, the attractors are formed reliably, and their population response is sustained robustly. Furthermore, the learning process is on-line and always on. Thanks to the spike-based Perceptron-like features of the learning circuits~\cite{Qiao_etal15,Brader_etal07}, the synapses stop increasing their weights once the attractors are fully formed, and can start to change again and adapt to the statistics of the input, should the input signals change. As discussed in Section~\ref{sec:attractor-networks}, these  attractor networks represent an extremely powerful computational primitive that can be used to implement sophisticated neuromorphic processing modules.}{}
In previous work~\cite{Indiveri_etal09,Chicca_etal14} we showed how the neuron, synapse, and learning circuits of the \ac{NP} can be used to implement \revised{othercp}{other types of powerful}{useful} computational primitives, such as the \ac{sWTA} network of Section~\ref{sec:learning}. In~\cite{Neftci_etal13} we showed how these primitives can be combined to synthesize fully autonomous neuromorphic cognitive agents able to carry out context-dependent decision making in an experiment analogous to the ones that are routinely done with primates to probe cognition. By properly defining the types of spiking neural networks in the \ac{NP}, and by properly setting its circuit parameters, it is therefore possible to build already now small scale embedded systems that can use its distributed memory resources to learn about the statistics of the input signals and of its internal state, while interacting with the environment in real time, and to provide state- and context-dependent information processing.
\revised{specialized}{}{To build more complex brain-like cognitive computing systems however it will be necessary to interface multiple neuromorphic devices or cores together with diverse and specialized functionalities, very much like the cortex uses multiple areas with different sensory, motor, and cognitive functional specifications~\cite{Douglas_Martin12}.}



\section{Emerging nano-technologies}
\label{sec:emerging}

An additional resource for building complex brain-like cognitive computing systems that are compact and low-power is provided by the large range of emerging  nano-scale devices that are being proposed to replace the functionality of larger and bulkier \ac{CMOS} circuits currently deployed for modeling synapses and neurons.
Recent research in nano-scale materials is revealing the possibility of using novel devices \revised{synapses}{to emulate the behavior of real synapses}{based on them to implement synapses} in artificial neural networks, and in particular to reproduce their learning and state-holding abilities. The general goal is to exploit the non-volatile memory properties of these devices and their ability to keep track of their state's past dynamics to implement massively parallel \revised{memcomputing}{arrays of nano-scale elements}{``memcomputing'' hardware~\cite{Di-Ventra_Pershin13}} integrated into neuromorphic \ac{VLSI} devices and systems. \revised{rolls-memristors1}{For example, in~\cite{Indiveri_etal13} we showed how it is possible to integrate memristive devices in \ac{CMOS} synapse arrays of the type used in the \ac{NP} of Section~\ref{sec:neur-proc}.}{}
A promising technology is the one of \acp{RRAM}~\cite{Yu_etal11a}, which exploit resistance switching phenomena~\cite{Linn_etal10}, and are very attractive due to their compatibility with \ac{CMOS} technology. \revised{rram}{The base element of a \ac{RRAM} device is a two-terminal element with a top electrode, a bottom one, and  a thin film, sandwiched between the electrodes. By applying a voltage across the electrodes, the electrical conductivity of the thin film material can be reversibly changed, from a high conductive to a high resistive state and viceversa,  and the corresponding conductance value can be stored for a long period.}{Indeed,} Several proposals have  been made for leveraging basic nano-scale \ac{RRAM} attributes in synapse circuits in neuromorphic architectures~\cite{Jo_etal10,Indiveri_etal13,Suri_etal13,Serrano-Gotarredona_etal13}; 
\revised{ram}{many of these proposals do not use these devices as conventional \ac{RAM} cells, but distribute them within and across  the synapse circuits in the neuromorphic architectures.}{}
It has been shown that these \ac{RRAM}-based neuromorphic approaches can potentially improve density and power consumption by at least a factor of 10, as compared with conventional \ac{CMOS} implementations~\cite{Rajendran_etal13}. 

Other approaches \revised{rram2}{that also store memory state as resistance, but that exhibit a range of different behaviors include}{alternative to resistive memories include the development of} \acp{STT-MRAM}~\cite{Chun_etal13,Vincent_etal14,Locatelli_etal14}, ferroelectric devices~\cite{Chanthbouala_etal12}, and phase change materials~\cite{Suri_etal12,Kuzum_etal12,Wimmer_Salinga14}. 
\revised{industrial}{In general, oxide-based \acp{RRAM}, \acp{STT-MRAM} and phase-change memories are under intense industrial development. Although, these technologies are currently difficult to access for non-industrial applications, basic research in this domain has very high potential, because}{mixed-signal and event-based} neuromorphic circuits can harness the interesting physics being discovered in these new devices to extend their applicability: in addition to developing nano-scale materials and devices that can emulate the biophysics of real synapses and neurons, this research can lead to understanding how to exploit their complex switching dynamics to reproduce relevant computational primitives, such as state-dependent conductance changes, multi-level stability and stochastic weight updates, for use in large-scale  neural processing systems~\cite{Indiveri_etal13,Saighi_etal15}.

\section{Discussion}
\label{sec:discussion}

\revised{discussion-largescale}{
The array of possible neuromorphic computing platforms described in this work, illustrates the current approaches used in tackling the partitioning of memory and information processing blocks on these systems. Here we discuss about the advantages and disadvantages of the different approaches and summarize the features of the neuromorphic computing platforms presented.

\paragraph{Large-scale simulation platforms}  
Neural network simulation frameworks based on C or Python~\cite{Brette_Goodman12,Brette_etal07,Richert_etal11} which can run on conventional \acp{CPU} and \acp{GPU} based systems, offer the best flexibility and quickest development times for simulating large-scale spiking networks. These simulations however can still take a long time on conventional computing platforms with the additional disadvantage of very high power consumption figures (e.g., up to tens of mega-Watts).  Dedicated hardware solutions have been built to support fast simulations of neural models, and to reduce their power-consumption figures. Hardware platforms such as SpiNNaker, BrainScales, and NeuroGrid fall under this category.

Questions still remain whether large-scale simulations are necessary to answer fundamental neuroscience questions~\cite{Markram06a,Fregnac_Laurent14}. It is also not clear whether the compromises and trade-offs made with  these custom hardware implementations will restrict the search of possible solutions to these fundamental neuroscience questions or dissuade neuroscientists from using such platforms for their work. For example, the models of neurons and synapses implemented on Neurogrid and BrainScales are hard-wired and cannot be changed. On the SpiNNaker platform they are programmable, but there are other constraints imposed on the type of neurons, synapses and networks that can be simulated by the limited memory, the limited resolution, and by the fixed-point representation of the system.

Another critical issue that affects all large-scale simulator platforms is the \ac{IO} bottleneck. Even if these hardware systems can simulate neural activity in real-time, or accelerated time (e.g.,  1\,ms of physical time simulated in 1\,$\mu$s), the time required to load the configuration and the parameters of a large-scale neural network can require minutes to hours: for example, even using the latest state-of-the-art technology transfer rates of 300\,Gb/s (e.g., with 12x EDR InfiniBand links), the time required to configure a single simulation run of a network comprising $10^6$ neurons with a fan-out of 1000, and a fan-in of 10000 synapses with 8-bit resolution weights would require at least 45 minutes.
}{}

\revised{truenorth}{\paragraph{General purpose computing platforms} In addition to dedicated simulation engines for neuroscience studies, neuromorphic information processing systems have also been proposed as general purpose non von Neumann computing engines for solving practical application problems, such as pattern recognition or classification.  Example platforms based on \ac{FPGA} and \ac{ASIC} designs using the standard logic design flow have been described in Section~\ref{sec:electronic-systems}.  Systems designed using less conventional design techniques or emerging nano-scale technologies  include the IBM TrueNorth system (see Section~\ref{sec:truenorth} for the former), and the  memristor- and \ac{RRAM}-based neuromorphic architectures (see Section~\ref{sec:emerging} for the latter). 
The TrueNorth system however does not implement learning and adaptation. Therefore it can only be used as a low-power neural computing engine once the values of the synaptic weights have been computed and uploaded to the network. The complex learning process that determines these synaptic weights is typically carried out on power-hungry standard- or super-computers. This rules out the possibility of using this and similar architectures in dynamic situations in which the system is required to adapt to the changes of the environment or of its input signals. Endowing these types of architectures with  learning mechanisms, e.g., using memristive or \ac{RRAM}-based devices, could lead to the development of non von Neumann  computing platforms that are more adaptive and general purpose. The state of development of these adaptation and learning mechanisms and of the nano-scale memristive technologies however is still in its early stages, and the problems related to the control of learning dynamics, stability, and variability are still an active area of research~\cite{Saighi_etal15,Prezioso_etal15}.}{}

\revised{smallscale}{\paragraph{Small scale special purpose neuromorphic systems} 
Animal brains are not general purpose computing platforms. They
are highly specialized structures that evolved 
to increase the chances of survival in hazardous environments with limited resources and varying conditions~\cite{Allman00}. They represent an ideal computing technology for implementing robust feature extraction, pattern recognition, associative learning, sequence learning, planning, decision making, and ultimately for generating behavior~\cite{Churchland_Sejnowski92}.
}{}
\revised{smallscale2}{The original neuromorphic engineering approach~\cite{Mead90,Douglas_etal95a} proposed to develop and use electronic systems precisely for this purpose: to build autonomous cognitive agents that produce behavior in response to multiple types of varying input signals and different internal states~\cite{Neftci_etal13,Sandamirskaya13}. As argued in Section~\ref{sec:introduction}, the best way to reach this goal is to use electronic circuits biased in the sub-threshold regime~\cite{Liu_etal02a,Chicca_etal14}, and to directly emulate the properties of real neural systems by exploiting the physics of the Silicon medium. Examples of systems that follow this approach are the ones described in Sections~\ref{sec:spinnaker} and~\ref{sec:neur-proc}. The types of signals that these systems are optimally suited to  process include multi-dimensional auditory and visual inputs, low-dimensional temperature and pressure signals,  bio-signals measured in living tissue, or even real-time streaming digital bit strings, e.g., obtained from internet, Wi-Fi, or telecommunication data. Examples of application domains that could best exploit the properties of these  neuromorphic systems include wearable personal assistants,  co-processors in embedded/mobile devices, intelligent brain-machine interfaces for prosthetic devices, and sensory-motor processing units in autonomous robotic platforms.
}{}
\revised{memory}{\paragraph{Memory and information processing} Classical von Neumann computing architectures face the von Neumann bottleneck problem~\cite{Backus78}. We showed in Section~\ref{sec:electronic-systems} how current attempts to reduce this problem, e.g. by introducing cache memory close to the \ac{CPU} or by using general purpose \acp{GPU}, are not viable, if energy consumption is factored in~\cite{Cassidy_Andreou12}.  We then described dedicated \ac{FPGA} and full custom \ac{ASIC} architectures that carefully balance the use of memory and information processing resources for implementing deep networks~\cite{Pham_etal12,Dundar_etal14,Conti_Benini15} or large-scale computational neuroscience models~\cite{Farahini_etal14}. While these dedicated architectures,  still based on frames or graded (non-spiking) neural network models, represent an improvement over \ac{CPU} and \ac{GPU} approaches, the event-based architectures described in Sections~\ref{sec:frame-based-solution} and~\ref{sec:fpga-dbns} improve access to cache memory structures even further, because of their better use of locality in both space \emph{and} time. 

Also the SpiNNaker system, described in Section~\ref{sec:spinnaker}, exploits event-based processing and communication to optimize the use of memory and computation resources: computation is carried out by the system's parallel \ac{ARM} cores, while memory resources have been carefully distributed within each core (e.g., for caching data and instructions), across in-package \ac{DRAM} memory-chips (e.g., for storing program variables encoding network parameters) and in routing tables (e.g., for storing and implementing the network connectivity patterns). SpiNNaker and the event-based architectures of Sections~\ref{sec:frame-based-solution} and~\ref{sec:fpga-dbns} however still separate, to a large extent, the computation from memory access, and implement them in physically different circuits and modules. Conversely, TrueNorth, NeuroGrid, BrainScales and \ac{NP} architectures described in Sections~\ref{sec:cust-memory-optim} and~\ref{sec:neur-proc} represent a radical departure from the classical von Neumann computer design style. In TrueNorth (Section~\ref{sec:truenorth}) for example, the synapses (i.e., memory elements) are physically adjacent to the neuron circuits (i.e., computation elements), and multiple neuro-synaptic cores are distributed across the chip surface. Synapses in this architecture are used as basic binary memory elements and computation is mostly relegated to the neuron circuits (memory and computation elements are distributed and physically close to each other, but not truly co-localized). As discussed in Section~\ref{sec:neurogrid}, the NeuroGrid architecture follows a substantially different approach. Rather than implementing binary memory circuits in each synapse, it uses circuits with global shared parameters that emulate the temporal dynamics of real synapses with biologically realistic time constants. Computation is therefore carried out in both neurons and synapses, and the main memory requirements for programming and re-configuring the network's topology and function are in the routing tables and in the shared global parameters. Since a large part of the network topology is hardwired, the brange of possible neural models and functions that can be best emulated by NeuroGrid is restricted (by design) to models of cortical structures (e.g. parts of visual cortex). The BrainScales project (Section~\ref{sec:brainscales}) on the other hand, aims to support the simulation of large-scale networks of arbitrary topology, and which include non-linear operations at the synapse level (e.g., such as spike-timing dependent plasticity). Therefore the memory structures that set the network parameters are truly distributed and co-located with the computational elements. These include floating gate devices that set neuron and synapse parameter values as well as \ac{SRAM} and \acp{DAC} circuits that store the synaptic weights. The BrainScales memory elements that are used to store the network connectivity patterns, on the other hand, are distributed across multiple devices, and interfaced to very fast routing circuits designed following the conventional digital communication approach~\cite{Schemmel_etal10}.
A compromise between the highly flexible re-configurable but energy consuming approach of BrainScales, and the ultra-low power but with restricted degree of configurability approach of NeuroGrid is the one followed with the \ac{NP} (Section~\ref{sec:neur-proc}). In this system all memory resources, both for  routing digital events and for storing synaptic and neural circuit parameters are tightly integrated with the synapse and neuron computing circuits. Since the memory of the events being processed is stored in the dynamics of the circuits, which  have time constants that are well matched to the type of computation being carried out (see also the NeuroGrid real-time arguments on page~\pageref{sec:neurogrid}), memory and computation are  co-localized. The strategies used by the \ac{NP} for implementing multiple types of memory structures analogous to those used in conventional von Neumann architectures are the ones summarized in Table~\ref{tab:memory}. For example, we showed in Section~\ref{sec:neur-proc} how to build associative memories, by training a network of plastic neurons to memorize different patterns in four different attractors. The long-term memory changes were made in the network's synaptic weights, via the chip's spike-timing based learning mechanisms. Bi-stable or short-term memory structures can be made using the learned network attractors, that represent state-holding elements which emulate working-memory structures in cortical circuits, and which can be employed for state-dependent computation, for example implementing neural analogs of \acp{FSM}~\cite{Neftci_etal13,Rutishauser_Douglas09}. Alternative training protocols and network connectivity patterns can be used in the same chip to  carry out different types of neural information processing tasks, such as  binary classification, e.g. for image recognition tasks~\cite{Qiao_etal15}. Admittedly, the \ac{NP} comprises fewer neurons than those implemented in the large-scale neural systems surveyed in Section~\ref{sec:cust-memory-optim}, so the problem of allocating on-chip memory resources for routing events and configuring different connectivity patterns is mitigated. 
To build more complex neuromorphic systems that can interact with the world in real-time and express cognitive abilities it will be necessary to consider more complex systems, which combine multiple neural information processing modules with diverse and specialized functionalities, very much like the cortex uses multiple areas with different sensory, motor, and cognitive functional specifications~\cite{Van-Essen_etal92,Douglas_Martin12}, and which restrict the possible connectivity patterns to a subset that maximizes functionality and minimizes routing memory usage, very much like cortex uses patchy connectivity patterns with sparse long-range connections and dense short-range ones~\cite{Muir_etal11,Markov_Kennedy13}. Within this context, the optimal neuromorphic processor would be a multi-core device in which each core could be implemented following different approaches, and in which the routing circuits and memory structures would be distributed (e.g., within and across cores), heterogeneous (e.g. using \ac{CAM}, \ac{SRAM}, and/or even memrsitive devices) and hierarchical (e.g., with intra-core level-one routers, inter-core level-two routers, inter-chip level-three routers, etc.)~\cite{Moradi_etal15}. The systems surveyed in this paper represent    sources of inspiration for choosing the  design styles of the the neural processing modules in each core and the memory structures to use, for different application areas.}{}

\section{Conclusions}
\label{sec:conclusions}

In this work we presented a survey of state-of-art neuromorphic systems and their usability for supporting deep network models,  cortical network models, and brain inspired cognitive architectures. We outlined the trade-offs that these systems face in terms of memory requirements, processing speed, bandwidth, and their ability to implement the different types of computational primitives found in biological neural systems. We presented a mixed signal analog/digital neuromorphic processor and discussed how that system, as well as analogous ones being developed by the international research community, can be used to implement cognitive computing. Finally, \revised{conclusion}{in the discussion section, we highlighted the advantages and disadvantages of the different approaches being pursued, pointing out their strengths and weaknesses. In particular, we argued that while there currently is a range of very interesting and promising neuromorphic information processing systems available, they do not yet provide substantial advantages over conventional computing architectures for large-scale simulations, nor are they complex enough to implement specialized small-scale cognitive agents that can interact autonomously in any environment.

The tremendous progress in micro-electronics and nano-technologies in recent years has been paralleled by remarkable progress in both experimental and theoretical neuroscience.
To obtain significant breakthroughs in neuromorphic computing systems
that can demonstrate the features of biological systems such as robustness, learning abilities and possibly cognitive abilities, continuing research and development efforts are required in this interdisciplinary approach which involves neuroscientists, computer scientists, technologists, and material scientists.
This can be achieved by training a new generation of researchers with interdisciplinary skills and by encouraging the communities specializing on these different disciplines to work together as closely as possible, as is currently being done in several computational neuroscience academic institutions, such as, for example, the Institute of Neuroinformatics - University of Zurich and ETH Zurich, or at neuroscience and neuromorphic engineering workshops, such as the Telluride and CapoCaccia Neuromorphic Engineering Workshops~\cite{CapoCaccia,Telluride}.}{}

\section*{Acknowledgments}
We are grateful for the reviewers of this manuscript for their constructive comments andhelpful suggestions.
Many of the concepts presented here were inspired by the discussions held at the CapoCaccia Cognitive Neuromorphic Engineering Workshop and by the work of Rodney Douglas, Misha Mahowald, Kevan Martin, Matthew Cook, and Stefano Fusi. We are particularly grateful to  Rodney Douglas and Kevan Martin for their support and for providing an outstanding intellectual environment at the Institute of Neuroinformatics in Zurich. The \ac{NP} was designed and characterized together with N. Qiao, H. Mostafa, F. Corradi, M. Osswald, F. Stefanini, and D. Sumislawska. The attractor network and associative memory experimental results presented in Section~\ref{sec:neur-proc} were obtained by Federico Corradi, at the 2015 CapoCaccia Workshop. This work was supported by the EU ERC Grant ``neuroP'' (257219), and by the EU FET Grant ``SI-CODE'' (284553) and ``RAMP'' (612058). S. Liu is supported in part by the Swiss National Foundation Stochastic Event Inference Processor Grant \#200021\_135066.


\bibliographystyle{IEEEtran}
\bibliography{biblio}

\end{document}